\pdfoutput=1

\documentclass[11pt]{article}

\usepackage[final]{acl}

\usepackage{times}
\usepackage{latexsym}

\usepackage[T1]{fontenc}

\usepackage[utf8]{inputenc}

\usepackage{microtype}

\usepackage{inconsolata}

\usepackage{graphicx}
\usepackage{dsfont}

\usepackage{enumitem}
\usepackage{rotating}

\usepackage{booktabs}
\usepackage{multirow}
\usepackage{tcolorbox}
\usepackage{subcaption}

\usepackage{arydshln}

\usepackage{amsthm}
\usepackage{amsmath}
\usepackage{amssymb}
\usepackage{xspace}
\newcommand{\narrowtextsc}[1]{\textls[-50]{\textsc{#1}}}
\newcommand{\lm}[1]{\texttt{#1}}
\newcommand{\sys}[1]{\narrowtextsc{#1}}
\newcommand{\data}[1]{\textsf{#1}}

\usepackage{xcolor,colortbl}

\usepackage{soul}
\definecolor{lightblue}{RGB}{158, 198, 243}
\definecolor{blue}{RGB}{224,230,255}
\definecolor{lighpurple}{RGB}{255, 241, 213}
\definecolor{purple}{RGB}{254,186, 251}
\definecolor{deepred}{RGB}{254,168,168}
\definecolor{deepgreen}{RGB}{170,229,181}
\definecolor{lightred}{RGB}{242,207, 194}
\definecolor{gray}{RGB}{216,216,216}
\definecolor{ensemble}{RGB}{223,168,254}

\usepackage{verbatim}

\title{Truth or Twist? Optimal Model Selection for Reliable Label Flipping Evaluation in LLM-based Counterfactuals}

\newcommand{\affilsup}[1]{\rlap{\textsuperscript{\normalfont#1}}}

\author{
    Qianli Wang\affilsup{1,3,\footnotemark[1]}
    \qquad 
    Van Bach Nguyen\affilsup{2,\footnotemark[1]}
    \quad
    Nils Feldhus\affilsup{1,3,5}
    \qquad
    \textbf{Luis Felipe Villa-Arenas\affilsup{1,3,4}}
    \\
    \textbf{Christin Seifert\affilsup{2}}
    \qquad
    \textbf{Sebastian M\"oller\affilsup{1,3}}
    \qquad
    \textbf{Vera Schmitt\affilsup{1,3}}
    \\
    $^1$Quality and Usability Lab, Technische Universit\"at Berlin
    \quad
    $^2$University of Marburg \\
    $^3$German Research Center for Artificial Intelligence (DFKI)
    \quad
    $^4$Deutsche Telekom\\
    $^5$BIFOLD – Berlin Institute for the Foundations of Learning and Data\\
    \small{\textbf{Correspondence}: 
  \texttt{\href{mailto:qianli.wang@tu-berlin.de}{qianli.wang@tu-berlin.de}}
  \qquad
  \texttt{\href{mailto:vanbach.nguyen@uni-marburg.de}{vanbach.nguyen@uni-marburg.de}}
  }
}

\begin{document}
\maketitle
\renewcommand{\thefootnote}{\fnsymbol{footnote}}
\footnotetext[1]{Equal Contribution.}
\renewcommand*{\thefootnote}{\arabic{footnote}}
\begin{abstract}
Counterfactual examples are widely employed to enhance the performance and robustness of large language models (LLMs) through counterfactual data augmentation (CDA). However, the selection of the judge model used to evaluate label flipping, the primary metric for assessing the validity of generated counterfactuals for CDA, yields inconsistent results. 
To decipher this, we define four types of relationships between the counterfactual generator and judge models: being the same model, belonging to the same model family, being independent models, and having an distillation relationship. Through extensive experiments involving two state-of-the-art LLM-based methods, three datasets, four generator models, and 15 judge models, complemented by a user study $(n = 90)$, we demonstrate that judge models with an independent, non-fine-tuned relationship to the generator model provide the most reliable label flipping evaluations.\footnote{Code and evaluation results are available at: \url{https://github.com/qiaw99/truth-or-twist}}
Relationships between the generator and judge models, which are closely aligned with the user study for CDA, result in better model performance and robustness. Nevertheless, we find that the gap between the most effective judge models and the results obtained from the user study remains considerably large. This suggests that a fully automated pipeline for CDA may be inadequate and requires human intervention.
 \looseness=-1
\end{abstract}

\section{Introduction}
\begin{figure*}[t]
\centering
\resizebox{0.93\textwidth}{!}{
\begin{minipage}{\columnwidth}
\includegraphics[width=\columnwidth]{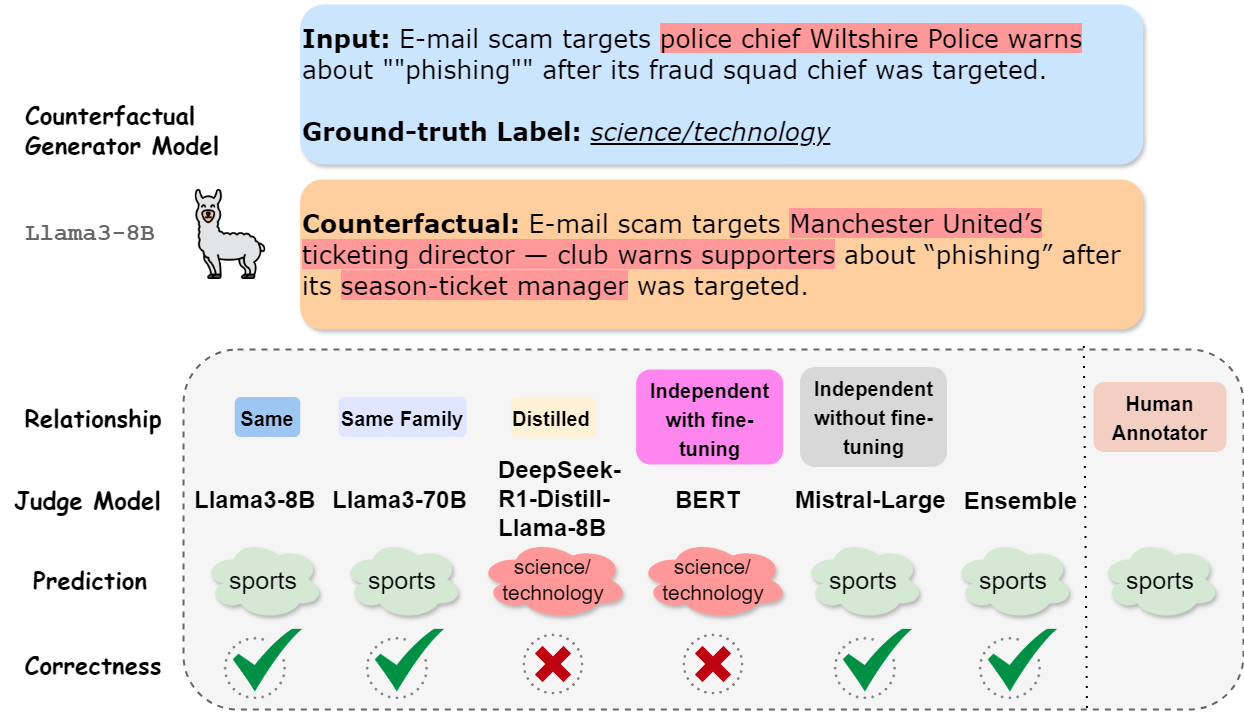}
\end{minipage}
}
\caption{A counterfactual generated by \lm{Llama3-8B}, with its label evaluated by judge models with different relationships, complemented by human evaluation. The revised words are highlighted in red.}
\label{fig:example}
\end{figure*}

Counterfactual examples are minimally altered versions of original inputs that flip the initial label \cite{miller-2019-explanation, ross-etal-2021-explaining, madsen-2022-survey}. They serve as a valuable approach for CDA aimed at improving model robustness and performance \cite{liu-etal-2021-counterfactual, dixit-etal-2022-core, balashankar-etal-2023-improving, agrawal2025enhancing}. We want to emphasize the subtle yet significant distinction between counterfactuals used for explaining model predictions and those used for CDA. In the former, the objective is to flip \textbf{the model's prediction}, while the goal of CDA is to flip \textbf{the ground truth label} (Figure~\ref{fig:app_example} in Appendix~\ref{app:datasets}).\footnote{In this study, we mainly focus on the latter type of counterfactuals used for CDA.} To evaluate the effectiveness and validity of the LLM-generated counterfactuals for CDA, the label flip rate (LFR) is a common metric of choice \cite{ge-2021-counterfactualevaluationexplainableai}. It is the percentage of valid counterfactuals where the ground truth labels are flipped out of the total number of instances. LFR of counterfactuals can be evaluated using either the same model that generates the counterfactual \cite{bhattacharjee-etal-2024-zero, bhattacharjee2024llmguidedcausalexplainabilityblackbox, wang2025fitcfframeworkautomaticfeature} or independent models \cite{dixit-etal-2022-core, balashankar-etal-2023-improving}. 
The optimal strategy for selecting models to evaluate the ground-truth validity of counterfactuals remains uncertain (Figure~\ref{fig:example}). This uncertainty, in turn, hampers efforts to enhance model robustness and performance through CDA, as noisy or erroneous labels may degrade model performance. 

In this work, we \textbf{first} define four types of relationships between the counterfactual generator model and the judge model: being the same model, independent models with and without fine-tuning on the target dataset, distilled models, and models from the same family (Figure~\ref{fig:example}). \textbf{Secondly}, we conduct comprehensive experiments to predict labels for counterfactuals generated by two state-of-the-art approaches across three datasets and four generator models, with 15 judge models. \textbf{Thirdly}, we undertake a user study to assess the validity of the generated counterfactuals in acquiring a ground-truth LFR. 

We find that a judge model with an independent, non-fine-tuned relationship to the generator captures label flipping most effectively. Relationships between the generator and judge models that are most aligned with the user study, lead to improved model performance and robustness. 
Additionally, there remains a considerable gap between the performance of the best judge models and the results observed in the user study.  \looseness=-1



\section{Background and Related Work}

\paragraph{CDA Pipeline} At the start of a CDA process, LLMs generate counterfactuals using established counterfactual generation approaches (Figure~\ref{fig:cda_pipeline} in Appendix~\ref{app:cda_pipeline}). These generated counterfactuals should subsequently be validated -- either by human annotators or judge models -- as invalid counterfactuals bearing incorrect labels may degrade model performance~\cite{PMID:35254993}. In practice, relying solely on human evaluation is both costly and inefficient; therefore, LLMs are commonly employed to validate the generated counterfactuals. Finally, valid counterfactuals are utilized as additional training data to enhance model performance and robustness \cite{yang-etal-2021-exploring, dixit-etal-2022-core}.

\paragraph{Model Selection for Label Flipping Evaluation}
The literature identifies two primary ways to verify label flipping in edited inputs: (1) employing an independent model, distinct from the one producing counterfactuals; (2) utilizing the same LLM that generates counterfactuals, guided by carefully constructed prompts. 
Prior to, and even during the widespread adoption of LLMs, encoder-only models, e.g., \lm{DeBERTa} \cite{dixit-etal-2022-core}, \lm{RoBERTa} \cite{ross-etal-2021-explaining, balashankar-etal-2023-improving, treviso-etal-2023-crest} or \lm{BERT} \cite{Kaushik2020Learning, fern-pope-2021-text,robeer-etal-2021-generating-realistic}, were predominantly used to verify label flipping. This preference stems from their superior performance in text classification tasks. In more recent work, the same LLMs are increasingly employed both to generate counterfactuals  \cite{bhattacharjee-etal-2024-zero, bhattacharjee2024llmguidedcausalexplainabilityblackbox, wang2025fitcfframeworkautomaticfeature, dehghanighobadi2025llmsexplaincounterfactually} and to evaluate them for label flipping, since their classification accuracy rivals that of fine-tuned encoder-only models. \looseness=-1

\section{Problem Framing}
\label{sec:problem}
\subsection{Label Flipping}
\label{subsec:lfr}
Given that counterfactual examples used for CDA are defined as edited inputs that alter \textit{ground truth} labels, LFR is positioned as the primary evaluation metric for assessing the effectiveness and validity of generated counterfactuals \cite{Kaushik2020Learning, dixit-etal-2022-core, zhu-etal-2023-explain, chen-etal-2023-disco}. LFR is quantified as the percentage of instances in which labels are successfully flipped relative to the total number of counterfactuals, where $N$ stands for the total number of counterfactuals, $y_{k}$ represents the ground-truth label of the original input, $y_{k}^{'}$ denotes the prediction of its corresponding counterfactual and $\mathds{1}$ is the indicator function: 
\begin{equation*}
    LFR = \frac{1}{N}\sum_{n=1}^{N} \mathds{1} (y_{k}^{'} \neq y_{k})
\end{equation*}

\subsection{Relationships}
\label{subsec:relationship}
To determine the optimal model selection strategy for label flip evaluation, we identified four prevalent relationships $\mathcal{R}(\mathcal{G},\mathcal{J})$ between the generator model $LLM_\mathcal{G}$ and judge models $LLM_\mathcal{J}$ (Figure~\ref{fig:example}), which are used to assess label flipping: 
\begin{itemize}[nosep,leftmargin=*,labelsep=0.4em,align=parleft,topsep=0pt,partopsep=0pt]
    \item {\setlength{\fboxsep}{0pt}\colorbox{lightblue}{\textbf{Same model} ($\mathcal{R}_{sm}$)}}: the two models are same. 
    \vspace{-0.5em}
    \begin{equation*}
        LLM_\mathcal{G} = LLM_\mathcal{J}
    \end{equation*}

    \item {\setlength{\fboxsep}{0pt}\colorbox{lighpurple}{\textbf{Same model family} ($\mathcal{R}_{sf}$)}}: the two models originate from the same model family $\mathcal{F(M)}$.
    \vspace{-0.5em}
    \begin{equation*}
        LLM_\mathcal{G}, LLM_\mathcal{J} \in \mathcal{F(M)}
    \end{equation*}

        \item \textbf{Independent models ($\mathcal{R}_{im}$)}: the two models belong to different model families. 
            \vspace{-0.5em}
    
    \begin{equation*}
        LLM_\mathcal{G} \in \mathcal{F}(\mathcal{M}_1), LLM_\mathcal{J} \in \mathcal{F}(\mathcal{M}_2)
    \end{equation*}
        We further distinguish the independent models based on whether $LLM_{\mathcal{J}}$ is fine-tuned on the given dataset: independent models {\setlength{\fboxsep}{0pt}\colorbox{purple}{\textit{with} ($\mathcal{R}_{imw}$)}} and {\setlength{\fboxsep}{0pt}\colorbox{gray}{\textit{without} ($\mathcal{R}_{imwo}$)}} fine-tuning.


    \item {\setlength{\fboxsep}{0pt}\colorbox{blue}{\textbf{Distilled models} ($\mathcal{R}_{dm}$)}}: $LLM_\mathcal{G}$ and $LLM_\mathcal{J}$ have an equal number of parameters and the same architecture. $LLM_\mathcal{J}$ is distilled and fine-tuned using synthetic data from a third model $LLM_\mathcal{D}$, which is more powerful and not part of the model family $\mathcal{F(M)}$ of the generator and judge model. 
    \vspace{-0.5em}
    \begin{align*}
        LLM_\mathcal{D} &\notin \mathcal{F(M)}\\
        \text{Size} (LLM_\mathcal{J}) &= \text{Size}(LLM_\mathcal{G}) \\
        \text{Archit.}(LLM_\mathcal{J}) &= \text{Archit.}(LLM_\mathcal{G}) \\
        LLM_\mathcal{J} &= \text{Inherit}(LLM_\mathcal{D})\\
        \{LLM_\mathcal{G}, LLM_\mathcal{J}\} &\cap LLM_\mathcal{D} = \varnothing
    \end{align*}

\end{itemize}

\section{Experimental Setup}
\subsection{Counterfactual Methods Selection}
\label{subsec:methods}
We select two state-of-the-art approaches based on LLMs that are shown to generate counterfactual examples efficiently and effectively: \sys{FIZLE} \cite{bhattacharjee-etal-2024-zero} and FLARE~\cite{bhattacharjee2024llmguidedcausalexplainabilityblackbox}. \sys{FIZLE} first prompts LLMs to identify key words within the input and then leverages these words to guide the generation of counterfactual examples. Meanwhile, FLARE generates counterfactuals by prompting LLMs in three steps: extracting latent features, identifying relevant words linked to those features, and modifying these words to produce counterfactual examples.

\subsection{Datasets}
\label{subsec:datasets}
We use three widely studied classification tasks for counterfactual generation in the literature\footnote{Examples of the dataset and the label distribution are included in Appendix~\ref{app:datasets}.}: \textit{news topic classification}, \textit{sentiment analysis}, and \textit{natural language inference}.

\paragraph{AG News} \cite{zhang-2015-agnews} is designed for news topic classification and comprises news articles categorized into four distinct topics: \textit{World}, \textit{Sports}, \textit{Business}, and \textit{Science/Technology}.

\paragraph{SST2} \cite{socher-etal-2013-recursive} serves as a popular dataset for sentiment analysis, sourced from movie reviews. It consists of reviews annotated with binary sentiment labels: \textit{positive} or \textit{negative}.


\paragraph{SNLI} \cite{bowman-etal-2015-large} is a dataset for natural language inference and contains premise–hypothesis pairs, annotated with one of three relational categories: \textit{entailment}, \textit{contradiction}, or \textit{neutral}.

\subsection{Models}
\label{subsec:model}
We select four LLMs varying in parameter size -- \lm{Qwen2.5}-\{14B,32B\} \cite{qwen2024qwen25technicalreport}, \lm{Llama3}-\{8B,70B\} \cite{llama3modelcard} --  to generate counterfactuals and serve as judge models $LLM_\mathcal{J}$ as {\setlength{\fboxsep}{0pt}\colorbox{lightblue}{$\mathcal{R}_{sm}$}} relationship (\S\ref{subsec:relationship}). 
Additionally, we deploy fine-tuned \lm{BERT} \cite{devlin-etal-2019-bert} and \lm{RoBERTa} \cite{liu2020roberta} on the target datasets (\S\ref{subsec:datasets}), along with off-the-shelf \lm{Phi4-14B} \cite{abdin2024phi4technicalreport}, \lm{Qwen2.5-72B} \cite{qwen2024qwen25technicalreport}, \lm{Mistral}-Large-Instruct \cite{jiang2023mistral7b}, DeepSeek-R1-Distill-\lm{\{Qwen,Llama\}} \cite{deepseekai2025deepseekr1incentivizingreasoningcapability}, and \lm{Gemini-1.5-pro} \cite{geminiteam2024gemini15unlockingmultimodal} as $LLM_\mathcal{J}$\footnote{Detailed information about the models employed is provided in Appendix~\ref{app:models}, and the downstream task performance of each model across three datasets and the classification prompts used are presented in Appendix~\ref{app:downstream_task}.} (Table~\ref{tab:automatic_evaluation}). We further ensemble label flipping results from all judge models via majority voting to yield final labels (\colorbox{ensemble}{\lm{ensemble}} in Figure~\ref{fig:mean_rank}). Moreover, since $LLM_\mathcal{J}$ are used to identify label flipping, we evaluate their downstream task performance in terms of \textbf{classification accuracy} across three datasets: $LLM_\mathcal{J}$ with {\setlength{\fboxsep}{0pt}\colorbox{purple}{$\mathcal{R}_{imw}$}} relationship (\lm{BERT} and \lm{RoBERTa}) generally achieve the highest downstream performance (Appendix~\ref{app:downstream_task}). 



\subsection{User Study}
\label{subsec:user_study}

We recruit 90 native English speakers and, for each of the three datasets, randomly sample 45 indices. For each subset, i.e., a generator-dataset pair (Table~\ref{tab:automatic_evaluation} in Appendix \ref{apdx:auto_eval}), the counterfactuals generated by the corresponding generator model $LLM_\mathcal{G}$ for the selected indices are evaluated by two human annotators. If no majority label emerges from the labels provided by human annotators, we break the tie ourselves by selecting one of the two annotated labels, ensuring the ground-truth label is agreed upon by two people.\footnote{The annotation guidelines and annotator information are provided in Appendix~\ref{app:annotation}. We further conduct an automatic evaluation in Appendix~\ref{app:sanity_check} on the selected 45 counterfactuals as a sanity check to validate their \textbf{representativeness of the overall distribution}.} Each annotator is given 15 counterfactuals, along with the set of possible labels given by the dataset, and tasked with selecting the optimal label. We report an inter-annotator agreement of Cohen's $\kappa=0.55$.\footnote{Although we employ two state-of-the-art methods -- FIZLE and FLARE -- to generate counterfactuals, they are not consistently perfect (at times failing to fully shift the semantics from the original to the target label) and sometimes produce ambiguous cases (Figure~\ref{fig:confusion}). This has an effect on the IAA being moderate which would likely improve with the development and use of more advanced counterfactual generation methods.} Lastly, we calculate the ground-truth LFR as the proportion of valid counterfactuals relative to the total number of instances (Table~\ref{tab:automatic_evaluation_values}  in Appendix~\ref{apdx:auto_eval}). \looseness=-1 

\subsection{Automatic Evaluation}
\subsubsection{Counterfactual LFR Evaluation} 
LFR is evaluated based on the classification results of counterfactual examples (\S\ref{subsec:lfr}) generated using \sys{FIZLE} and FLARE (\S\ref{subsec:methods}), using the deployed judge models $LLM_\mathcal{J}$ described in Section~\ref{subsec:model}.

\subsubsection{Human Alignment Evaluation}
To assess the alignment of $LLM_{\mathcal{J}}$ with human annotators, we employ three measures: (1) the average ranking ($rank \downarrow$, Figure~\ref{fig:mean_rank}); (2) the ratio of most-to-least alignment ($r_{m/l}\uparrow$); (3) Pearson correlation $\rho$ between human evaluation results and LFR results by judge models.

\paragraph{Average Ranking} To obtain the rankings, we first calculate LFR for human annotators and for each judge-generator model pair on each set of counterfactuals generated by given generator models, as reported in Table~\ref{tab:automatic_evaluation_values} of Appendix~\ref{apdx:auto_eval}. Next, we compute the difference ($\Delta$) between the human LFR and the LFR of each pair. A smaller difference indicates a better ranking (lower ranking value). This process results in a ranking for each judge-generator pair. Since these pairs correspond to specific relationships $\mathcal{R}(\mathcal{G},\mathcal{J})$, we average the rankings of all pairs sharing the same relationship to obtain the average ranking per relationship, as shown in Table~\ref{tab:automatic_evaluation} of Appendix~\ref{apdx:auto_eval}. Finally, we average these rankings for each generator model to produce the overall average rankings presented in Figure~\ref{fig:mean_rank}.

\paragraph{Most-to-Least-Alignment} Instead of measuring overall alignment, $r_{m/\ell}$  reports how many times each relationship $\mathcal{R}_{{\mathcal{G}, \mathcal{J}}}$ most or least closely aligned with human annotators across the three datasets:

\begin{align*}
    r_{m/\ell}(\mathcal{R}) = \frac{1}{|\mathcal{D}|} \sum_{d \in \mathcal{D}}\frac{\mathcal{N}_{\text{max}}(\mathcal{R},d)}{\mathcal{N}_{\text{min}}(\mathcal{R},d)}
\end{align*}

\noindent where $\mathcal{D}$ is the set of datasets and $\mathcal{N}_{\text{max}}$ and $\mathcal{N}_{\text{min}
}$ denote the number of cases, in which a generator-judge model pair is the most closely and least aligned with human evaluation results, respectively. A higher $r_{m/l}$ reflects better alignment.

\section{Results}
\label{sec:results}

\begin{figure}[t]
\centering
\resizebox{\columnwidth}{!}{
\begin{minipage}{\columnwidth}
\includegraphics[width=\columnwidth]{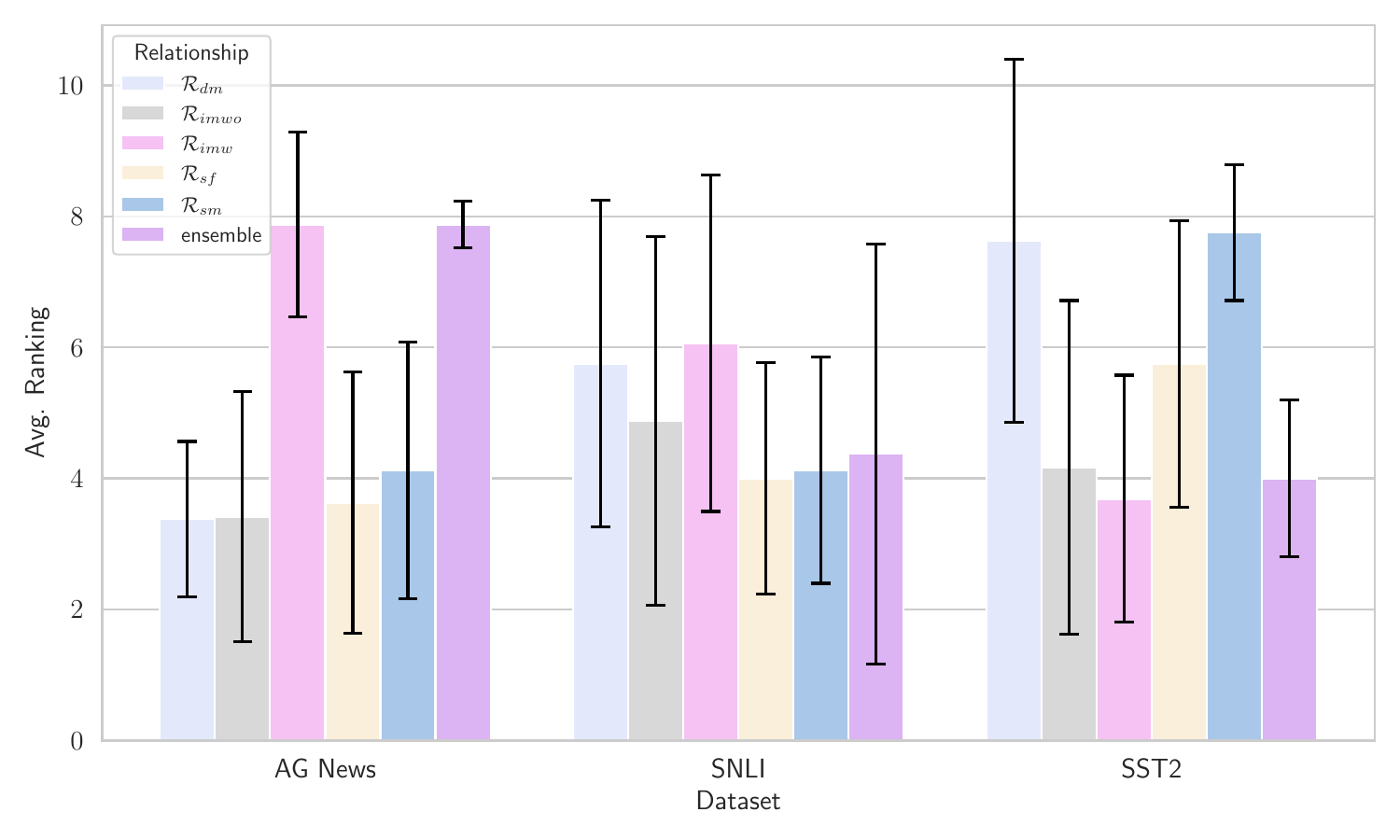}
\end{minipage}
}
\caption{The average ranking of judge-generator model relationship based on $\Delta$ from Table~\ref{tab:automatic_evaluation_values} in Appendix~\ref{apdx:auto_eval} (lower rankings indicate better alignment). Counterfactual examples are generated by \lm{Qwen2.5-\{14B,32B\}} and \lm{Llama3-\{8B,70B\}}, evaluated across judge models exhibiting \colorbox{lightblue}{same}, \colorbox{blue}{distilled}, \colorbox{lighpurple}{same family}, \colorbox{ensemble}{ensemble}, and independent \colorbox{purple}{w/} and \colorbox{gray}{w/o} fine-tuning relationships on \data{AG News}, \data{SNLI} and \data{SST2}.\looseness=-1}
\label{fig:mean_rank}
\end{figure}


\paragraph{Judge model performance depends on its relationship to the generator. } 

As shown in Figure~\ref{subfig:average_ranking} and Figure~\ref{subfig:most_to_least}, $LLM_{\mathcal{J}}$ with {\setlength{\fboxsep}{0pt}\colorbox{gray}{$\mathcal{R}_{imwo}$} relationship achieves the highest alignment with human annotators ($rank=4.15$, $r_{m/l}=3.5$, $\rho=0.47$). In contrast,  {\setlength{\fboxsep}{0pt}\colorbox{lightblue}{$\mathcal{R}_{dm}$}} ($rank=5.58$, $r_{m/l}=1$, $\rho=0.38$) or {\setlength{\fboxsep}{0pt}\colorbox{purple}{$\mathcal{R}_{imw}$} ($rank=5.86$, $r_{m/l}=0.23$, $\rho=0.29$) demonstrate poor alignment with human judgments. This can be attributed to data contamination \cite{li2025preferenceleakagecontaminationproblem}, as these models are either fine-tuned on the target dataset or share architectural similarities with $LLM_{\mathcal{G}}$. Such overlap could bias $LLM_{\mathcal{J}}$ in its evaluation of label flipping, potentially leading to either overestimating or underestimating the LFR. Notably, ensembling results from all judge models does not necessarily lead to better alignment, as it partially relies on results from suboptimal judge models. 
Additionally, we find that the first two measures (average ranking and $r_{m/\ell}$) are moderately correlated, with a Spearman correlation coefficient of 0.67, indicating that $r_{m/\ell}$ also serves as a reliable metric to capture the alignment between model pairs and human judgments. 


\begin{figure}[t!]
  \centering
  \begin{subfigure}{\columnwidth}
\centering
\includegraphics[width=\linewidth]{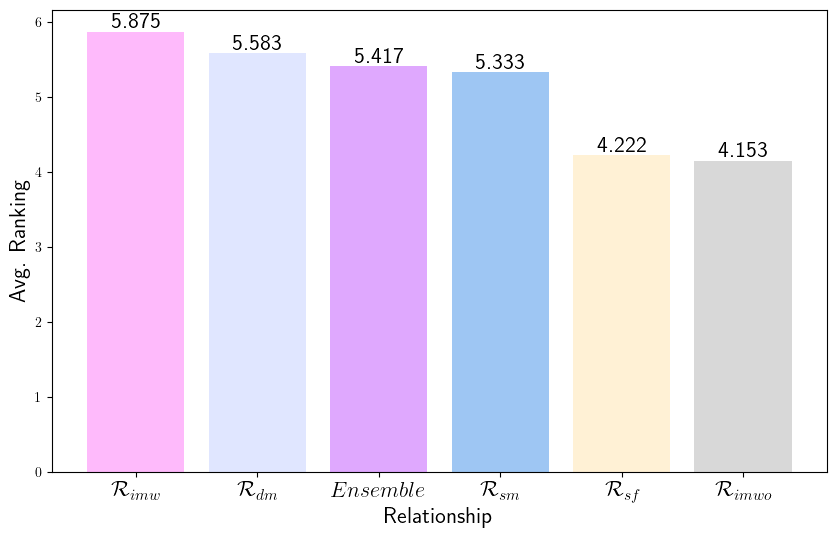}
\caption{Average ranking of the relationships $\mathcal{R}_{\mathcal{G}, \mathcal{J}}$.}
\label{subfig:average_ranking}

  \end{subfigure}
    \hfill
  \begin{subfigure}{\columnwidth}
\centering
\includegraphics[width=\linewidth]{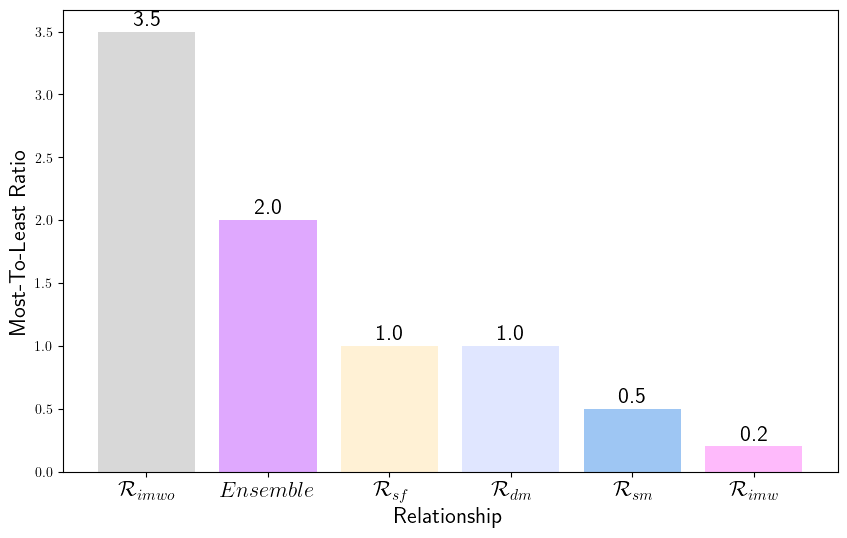}
\caption{Most-to-least ratio ($r_{m/\ell}$) of the relationships $\mathcal{R}_{\mathcal{G}, \mathcal{J}}$.}
\label{subfig:most_to_least}

  \end{subfigure}

  \caption{Average ranking and most-to-least ratio for the relationships $\mathcal{R}_{\mathcal{G}, \mathcal{J}}$: \colorbox{lightblue}{same}, \colorbox{blue}{distilled}, \colorbox{lighpurple}{same family}, \colorbox{ensemble}{ensemble}, and independent \colorbox{purple}{w/} and \colorbox{gray}{w/o} fine-tuning.}
  \label{fig:figure}
\end{figure}

\paragraph{High downstream performance does not necessarily indicate an effective judge model.}
Despite $LLM_{\mathcal{J}}$ with {\setlength{\fboxsep}{0pt}\colorbox{blue}{$\mathcal{R}_{dm}$}} or {\setlength{\fboxsep}{0pt}\colorbox{purple}{$\mathcal{R}_{imw}$}} relationships achieving the highest downstream task performance across the target datasets (Table~\ref{tab:performance} in Appendix~\ref{app:downstream_task}), their alignment with human annotators in evaluating label flipping remains considerably weak (Figure~\ref{fig:mean_rank}, Appendix~\ref{apdx:auto_eval}). This contrast highlights that strong downstream task performance does not lead to reliable label flipping evaluation, and it underscores the need for careful judge model selection. 

\paragraph{Identifying label flipping remains highly challenging.} LLMs may struggle to capture nuanced changes when determining the label of imperfect counterfactuals, as the context may not have fully shifted to support a different label, resulting in ambiguity (Figure~\ref{fig:example}, Appendix~\ref{app:challenges}). Notably, even the optimal judge model fails to fully capture label flipping, exhibiting an average discrepancy of 22.78\% relative to the user study results (Table~\ref{tab:automatic_evaluation_values} in Appendix \ref{apdx:auto_eval}), which implies that a fully automated CDA pipeline is insufficient and necessitates human oversight. 
 
\paragraph{Relationship $\mathcal{R}(\mathcal{G},\mathcal{J})$ impacts CDA outcomes.} We investigate how the choice of relationship affects CDA outcomes (Appendix~\ref{apdx:CDA}). We notice that when the LFR of $LLM_\mathcal{J}$, due to its relationship to $LLM_{\mathcal{G}}$, aligns more closely with user study outcomes, the labels identified by $LLM_\mathcal{J}$ are associated with improved performance and greater robustness on both unseen and out-of-distribution data \cite{Kaushik2020Learning,gardner2020evaluating} (Appendix~\ref{apdx:CDA}). This association is particularly evident when these identified labels are treated as the ground-truth labels for counterfactuals, which subsequently serve as data points for CDA. This can be attributed to the fact that noisy and incorrect labels provided by judge models with suboptimal relationships contribute to performance deterioration \cite{zhu-etal-2022-bert, PMID:35254993}.

\section{Conclusion}

In this work, we emphasize the importance of the relationship between the counterfactual generator model and label flipping judge model in achieving LFR that align more closely with human annotations and in improving CDA outcomes. We further demonstrate that high downstream performance does not necessarily imply an effective judge model. Through extensive experiments, we identify that label flipping remains highly challenging across all selected tasks. Additionally, the gap between the optimal relationship and the user study is considerably large, which indicates full automation of CDA falls short and human intervention should be considered.

\section*{Limitations}

Our experimental work is confined to English-language datasets. Consequently, the effectiveness in other languages may not be comparable. Extending experiments to the multilingual setting is considered as future work.

In our experiments, we exclusively use models from \lm{Qwen} and \lm{Llama} families to generate counterfactuals, as from \lm{DeepSeek-R1} \cite{deepseekai2025deepseekr1incentivizingreasoningcapability} distilled \lm{Qwen2.5} and \lm{Llama3} models are officially provided\footnote{\url{https://huggingface.co/collections/deepseek-ai/deepseek-r1-678e1e131c0169c0bc89728d}} and can be used out-of-the-box. Additional work is required when employing models from a different model family as $LLM_{\mathcal{G}}$, including using \lm{DeepSeek-R1} to generate synthetic data and fine-tuning $LLM_{\mathcal{G}}$ to derive $LLM_{\mathcal{J}}$ with distillation relationships. Between the model families (Qwen, Llama, Mistral, DeepSeek), there are lots of architectural equivalences and similarities, e.g., the same attention (grouped-query attention), position embeddings (RoPE), normalization (RMSNorm) or FFN activation (SwiGLU). We argue that, based on our comprehensive experiments and large-scale user study ($n=90$), our results are considerably robust and generalizable given the similar architectures compared to other model families. 

For label flipping evaluation, we select \lm{BERT}, \lm{RoBERTa}, \lm{Phi4-14B}, \lm{Mistral-Large-Instruct-2411} and \lm{Gemini-1.5-pro} as representatives of \textit{independent} ({\setlength{\fboxsep}{0pt}\colorbox{purple}{$\mathcal{R}_{imw}$}, \S\ref{subsec:relationship}) LLMs with different parameter sizes, without comprehensively assessing models from all other widely known model families. In particular, we evaluate only open-source models, rather than closed-source, proprietary models. \looseness=-1

Beyond LFR, counterfactuals can be evaluated subjectively -- via human judgment or LLM-as-a-Judge -- along dimensions such as \textbf{coherence}, \textbf{understandability}, \textbf{feasibility}, \textbf{fairness}, and \textbf{completeness} \cite{nguyen-etal-2024-ceval-benchmark, domnich2024unifyingevaluationcounterfactualexplanations, wang2025compressedlensinvestigatingimpact}. While automated metrics exist for other aspects (e.g., similarity, diversity), in this paper we focus on identifying which generator–judge relationship is preferable for verifying label flipping, as informed by our user-study results.

\section*{Ethics Statement}
The participants in our user studies were compensated at or above the minimum wage
in accordance with the standards of our host institutions’
regions. The annotation took each annotator 45
minutes on average.


\section*{Acknowledgment} 
We thank Leonhard Hennig for his review of an earlier paper draft of our paper. We are indebted to the anonymous reviewers of INLG 2025 for their helpful and rigorous feedback.
This work has been supported by the Federal Ministry of Research, Technology and Space (BMFTR) as part of the projects newspolygraph (03RU2U151C), BIFOLD 24B and VERANDA (16KIS2047).

\bibliography{custom}

\appendix

\section{Counterfactual Data Augmentation Pipeline}
\label{app:cda_pipeline}
Figure~\ref{fig:cda_pipeline} illustrates the CDA pipeline.
LLMs first generate counterfactuals using counterfactual generation approaches such as FIZLE and FLARE, both of which are used in our work (\S\ref{subsec:methods}). The generated counterfactuals must then be validated—either by human annotators or judge models—as invalid counterfactuals with incorrect labels may degrade model performance \cite{zhu-etal-2022-bert, PMID:35254993}. Finally, valid counterfactuals are utilized as additional training data to enhance model performance and robustness.

\begin{figure*}[t]
\centering
\resizebox{\textwidth}{!}{
\begin{minipage}{\columnwidth}
\includegraphics[width=\columnwidth]{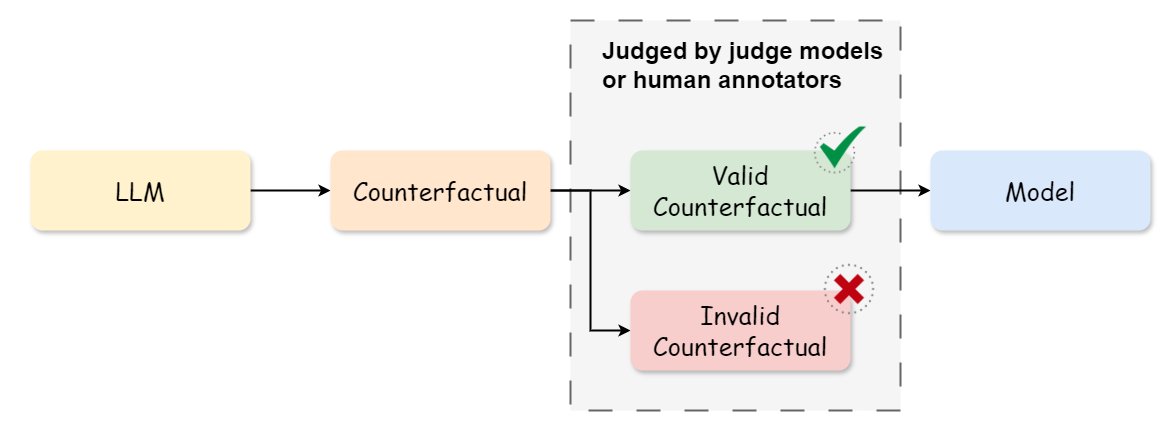}
\end{minipage}
}
\caption{Counterfactual Data Augmentation pipeline.}
\label{fig:cda_pipeline}
\end{figure*}

\section{Datasets}
\label{app:datasets}

\subsection{Label Distribution}
Figure~\ref{fig:app_label_distribution} shows the label distributions of \data{AG News}, \data{SNLI} and \data{SST2}.

\begin{figure}[t!]
  \centering

  \begin{subfigure}{\columnwidth}
    \centering
    \centering
\includegraphics[width=\linewidth]{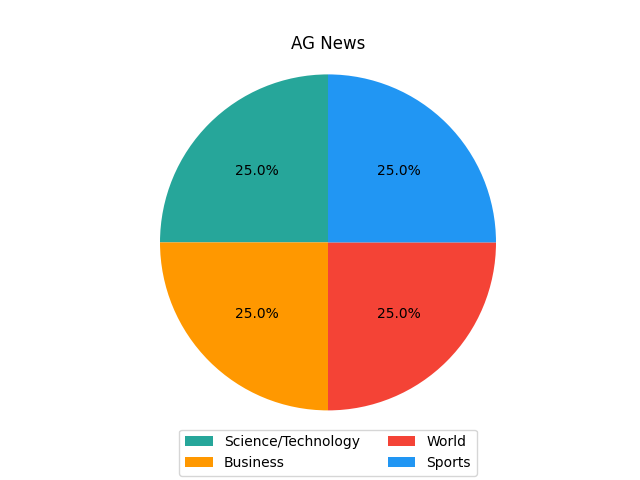}
\caption{\data{AG News}}
\label{fig:ag_news_distribution}
  \end{subfigure}
  \hfill
  \begin{subfigure}{\columnwidth}
\centering
\includegraphics[width=\linewidth]{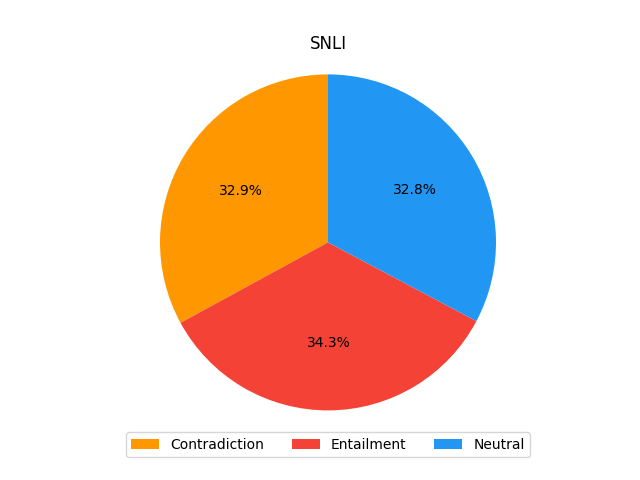}
\caption{\data{SNLI}}
\label{fig:snli_label_distribution}

  \end{subfigure}
    \hfill
  \begin{subfigure}{\columnwidth}
\centering
\includegraphics[width=\linewidth]{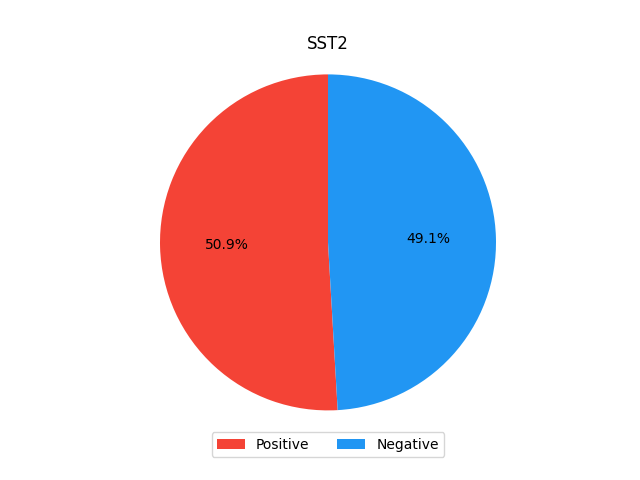}
\caption{\data{SST2}}
\label{fig:sst2_label_distribution}

  \end{subfigure}

  \caption{Label distributions of \data{AG News}, \data{SNLI} and \data{SST2}.}
  \label{fig:app_label_distribution}
\end{figure}


\subsection{Dataset Example}

Figure~\ref{fig:app_example} displays dataset examples from \data{AG News}, \data{SST2} and \data{SNLI}.

\begin{figure*}[h!]
    
    \centering

    \begin{tcolorbox}[colback=blue!30!white, colframe=black!10!blue, title=AG News (News Topic Classification)]
    \textbf{\textit{News:}} E-mail scam targets police chief Wiltshire Police warns about ""phishing"" after its fraud squad chief was targeted.

    \bigskip

    \textbf{\textit{Ground-truth Label:}} sci/tech

    \bigskip

    \textbf{Counterfactual:} E-mail scam targets \underline{Manchester United’s ticketing director} — \underline{club warns supporters} about “phishing” after its \underline{season-ticket manager} was targeted.

    \bigskip

    \textbf{Counterfactual label:} sports
    
    \end{tcolorbox}

    \begin{tcolorbox}[colback=blue!30!white, colframe=black!10!blue, title=SST2 (Sentiment Analysis)]
    \textbf{\textit{Review:}} Allows us to hope that nolan is poised to embark a major career as a commercial yet inventive filmmaker.

    \bigskip

    \textbf{\textit{Ground-truth Label:}} positive

    \bigskip

    \textbf{Counterfactual:} \underline{Dashes} any hope that \underline{N}olan is poised to embark on a major career as a commercial yet inventive filmmaker.

    \bigskip

    \textbf{Counterfactual label:} negative
    
    \end{tcolorbox}

    \begin{tcolorbox}[colback=blue!30!white, colframe=black!10!blue, title=SNLI (Natural Language Inference)]
    \textbf{\textit{Premise:}} This church choir sings to the masses as they sing joyous songs from the book at a church.
    
    \bigskip

    \textbf{\textit{Hypothesis:}} The church has cracks in the ceiling.
    
    \bigskip

    \textbf{\textit{Ground-truth Label:}} neutral

    \bigskip

    \textbf{Counterfactual (Hypothesis):} The church is \underline{completely empty and silent}.

    \bigskip

    \textbf{Counterfactual label:} contradiction
    \end{tcolorbox}

    \caption{Dataset Examples from \data{AG News}, \data{SST2} and \data{SNLI}.
    }
    \label{fig:app_example}
\end{figure*}

\section{Models}
\label{app:models}
Table~\ref{tab:used_model} provides detailed information about deployed models in our experiments. All models were directly obtained from the Hugging Face\footnote{\url{https://huggingface.co/}} repository. All experiments were conducted using A100 or H100 GPUs. For each model, counterfactual example generation across the entire dataset can be completed within 10 hours.
\begin{table*}[t!]
    \centering
    \resizebox{\textwidth}{!}{%
        \begin{tabular}{cccc}

        \toprule
        \textbf{Name}& \textbf{Citation} & \textbf{Size} & \textbf{Link}\\

        \midrule
        \lm{BERT} (\data{AG News}) & \citet{devlin-etal-2019-bert} & 110M & \url{https://huggingface.co/textattack/bert-base-uncased-ag-news}\\

        \lm{BERT} (\data{SST2}) & \citet{devlin-etal-2019-bert} & 110M & \url{https://huggingface.co/textattack/bert-base-uncased-SST-2} \\
        
        \lm{BERT} (\data{SNLI}) & \citet{devlin-etal-2019-bert} & 110M & 
        \url{https://huggingface.co/textattack/bert-base-uncased-snli} \\

        \lm{RoBERTa} (\data{AG News}) & \citet{liu2020roberta} & 125M & \url{https://huggingface.co/textattack/roberta-base-ag-news}\\

        \lm{RoBERTa} (\data{SST2}) & \citet{liu2020roberta} & 125M & \url{https://huggingface.co/textattack/roberta-base-SST-2}\\

        \lm{RoBERTa} (\data{SNLI}) & \citet{liu2020roberta} & 125M &
        \url{https://huggingface.co/pepa/roberta-base-snli}\\
        
        \lm{Llama3} & \citet{llama3modelcard} & 8B &   \url{https://huggingface.co/meta-llama/Meta-Llama-3-8B-Instruct}\\

        \lm{Llama3} & \citet{llama3modelcard} & 70B &   \url{https://huggingface.co/meta-llama/Meta-Llama-3-70B-Instruct}\\

        \lm{Qwen2.5} & \citet{qwen2024qwen25technicalreport} & 7B &   \url{https://huggingface.co/Qwen/Qwen2.5-7B-Instruct}\\
        
        \lm{Qwen2.5} & \citet{qwen2024qwen25technicalreport} & 14B &   \url{https://huggingface.co/Qwen/Qwen2.5-14B-Instruct}\\
        
        \lm{Qwen2.5} & \citet{qwen2024qwen25technicalreport} & 32B &   \url{https://huggingface.co/Qwen/Qwen2.5-32B-Instruct}\\
        
        \lm{Qwen2.5} & \citet{qwen2024qwen25technicalreport} & 72B &   \url{https://huggingface.co/Qwen/Qwen2.5-72B-Instruct}\\

        \lm{Phi4} &  \citet{abdin2024phi4technicalreport} & 14B & \url{https://huggingface.co/microsoft/phi-4}\\

        \lm{Mistral-Large-Instruct-2311} & \citet{jiang2023mistral7b} & 123B & \url{https://huggingface.co/mistralai/Mistral-Large-Instruct-2411}\\ 

        \lm{Gemini-1.5-pro} & \citet{geminiteam2024gemini15unlockingmultimodal} & n.a. & \url{https://gemini.google.com/}\\


        \lm{DeepSeek-R1-Distill-Qwen-14B} & \citet{deepseekai2025deepseekr1incentivizingreasoningcapability} & 14B & \url{https://huggingface.co/deepseek-ai/DeepSeek-R1-Distill-Qwen-14B} \\

        \lm{DeepSeek-R1-Distill-Qwen-32B} & \citet{deepseekai2025deepseekr1incentivizingreasoningcapability} & 32B & \url{https://huggingface.co/deepseek-ai/DeepSeek-R1-Distill-Qwen-32B} \\

        \lm{DeepSeek-R1-Distill-Llama-8B} & \citet{deepseekai2025deepseekr1incentivizingreasoningcapability} & 8B & \url{https://huggingface.co/deepseek-ai/DeepSeek-R1-Distill-Llama-8B}\\

        \lm{DeepSeek-R1-Distill-Llama-70B} & \citet{deepseekai2025deepseekr1incentivizingreasoningcapability} & 70B & \url{https://huggingface.co/deepseek-ai/DeepSeek-R1-Distill-Llama-70B}\\

        \bottomrule
        \end{tabular}
        }
    \caption{
    Detailed information about used models in our experiments. 
    }
    \label{tab:used_model}
\end{table*}

\section{Downstream Task Performance}
\label{app:downstream_task}
\begin{table*}[t!]
    \centering
    \setlength{\extrarowheight}{3pt}
    \renewcommand*{\arraystretch}{0.7}
    
    \footnotesize
    \resizebox{.65\textwidth}{!}{%
        \begin{tabular}{cccc}

        \toprule[1.5pt]
        Model & \data{AG News}  & \data{SST-2}  & \data{SNLI} \\ 
        
        \hline
        \lm{Qwen2.5-7B} & 78.93 & 93.23 & 88.07\\
        \lm{Qwen2.5-14B} & 82.80 & 93.23 &  82.43\\
        \lm{Qwen2.5-32B} & 81.79 & 94.50 & 85.67\\
        \lm{Qwen2.5-72B} & 78.30 & 94.38 & 85.60\\
        \lm{Llama3-8B} & 71.00 & 86.70 & 50.93\\
        \lm{Llama3-70B} & 83.10 & 94.15 & 62.70\\
        \hline
    
        \lm{DeepSeek-R1-Distill-Qwen-7B} & 76.92 & 78.56 &  58.87\\
        \lm{DeepSeek-R1-Distill-Qwen-14B} & 81.95 & 88.90 & 74.33 \\
        \lm{DeepSeek-R1-Distill-Qwen-32B} & 83.81 & 93.25 & 79.10\\
        \lm{DeepSeek-R1-Distill-Llama-8B} & 80.71 & 85.21 & 49.60\\
        \lm{DeepSeek-R1-Distill-Llama-70B} & 84.81 & 92.15 & 73.02\\
        \hline
        \lm{bert-base-uncased} & 95.14 & 92.32 & 87.50 \\
        \lm{roberta-base-uncased} & 94.69 & 94.04 & 88.60\\
        \lm{Phi4-14B} & 80.49 & 92.78 & 82.93\\
        \lm{Mistral-Large} & 79.93 & 84.40 & 85.73\\
        \lm{Gemini-1.5-pro} & 83.60 & 95.40 & 77.80\\
        
        \toprule[1.5pt]     
        
        \end{tabular}
    }
    \caption{Downstream task performance, qualified by $F_1$ score (in \%) on the \data{AG News}, \data{SST2} and \data{SNLI} datasets.
    }
    \label{tab:performance}
\end{table*}

Table~\ref{tab:performance} reports the downstream task performance for all LLMs presented in Table~\ref{tab:automatic_evaluation} and Section~\ref{subsec:model} on the \data{AG News}, \data{SST2}, and \data{SNLI} datasets. Zero-shot prompting is applied to \lm{DeepSeek-R1-Distill-\{Qwen,Llama\}}, as few-shot prompting consistently impairs their performance \cite{deepseekai2025deepseekr1incentivizingreasoningcapability}. Furthermore, as observed in Figure~\ref{fig:ag_news_performance} and Figure~\ref{fig:snli_performance}, there is an inverse correlation between the number of demonstrations and classification accuracy for the \data{AG News} and \data{SNLI} datasets, aligned with the finding of \citet{vajjala2025textclassificationllmera}. Similarly, in Figure~\ref{fig:sst2_performance}, increasing number of demonstrations does not yield significant benefits and, in some cases, even degrades performance for the \data{SST2} dataset. Therefore, zero-shot prompting is employed for all other decoder-only LLMs as well. The used prompt instructions are shown in Figure~\ref{fig:classification_prompt}.

\begin{figure*}[t!]
\centering
\resizebox{\textwidth}{!}{
\begin{minipage}{\columnwidth}
\includegraphics[width=\columnwidth]{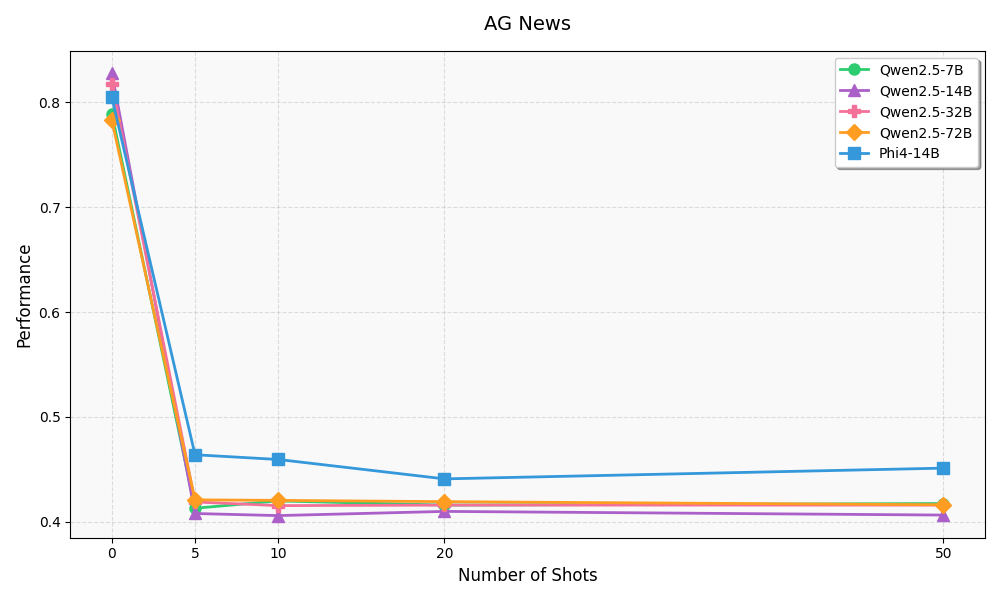}
\end{minipage}
}
\caption{Classification performance of models on the \data{AG News} dataset under different few-shot learning scenarios ($n \in \{0,5,10,20,50\}$).}
\label{fig:ag_news_performance}
\end{figure*}

\begin{figure*}[t!]
\centering
\resizebox{\textwidth}{!}{
\begin{minipage}{\columnwidth}
\includegraphics[width=\columnwidth]{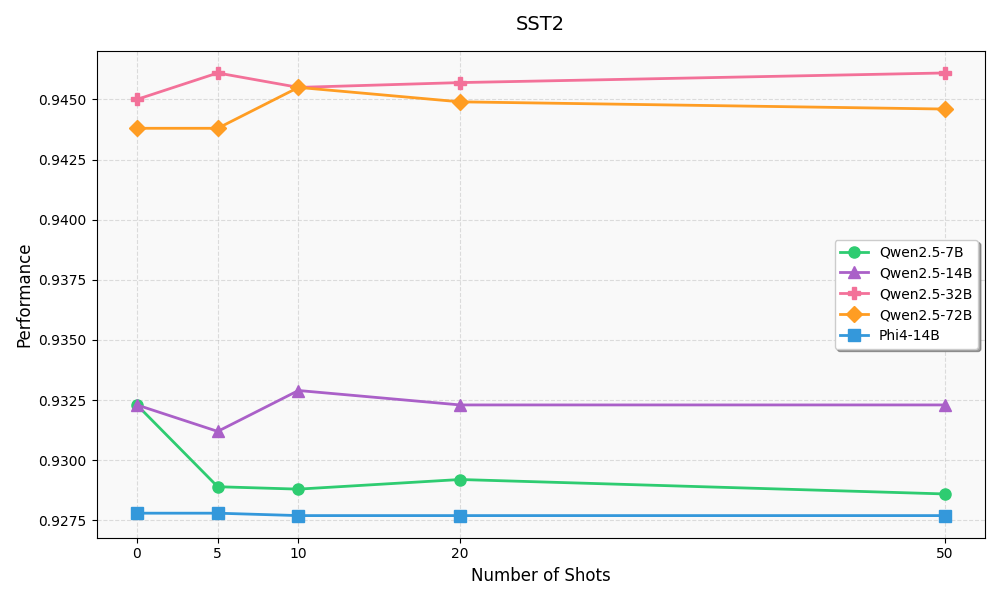}
\end{minipage}
}
\caption{Classification performance of models on the \data{SST2} dataset under different few-shot learning scenarios ($n \in \{0,5,10,20,50\}$).}
\label{fig:sst2_performance}
\end{figure*}

\begin{figure*}[t!]
\centering
\resizebox{\textwidth}{!}{
\begin{minipage}{\columnwidth}
\includegraphics[width=\columnwidth]{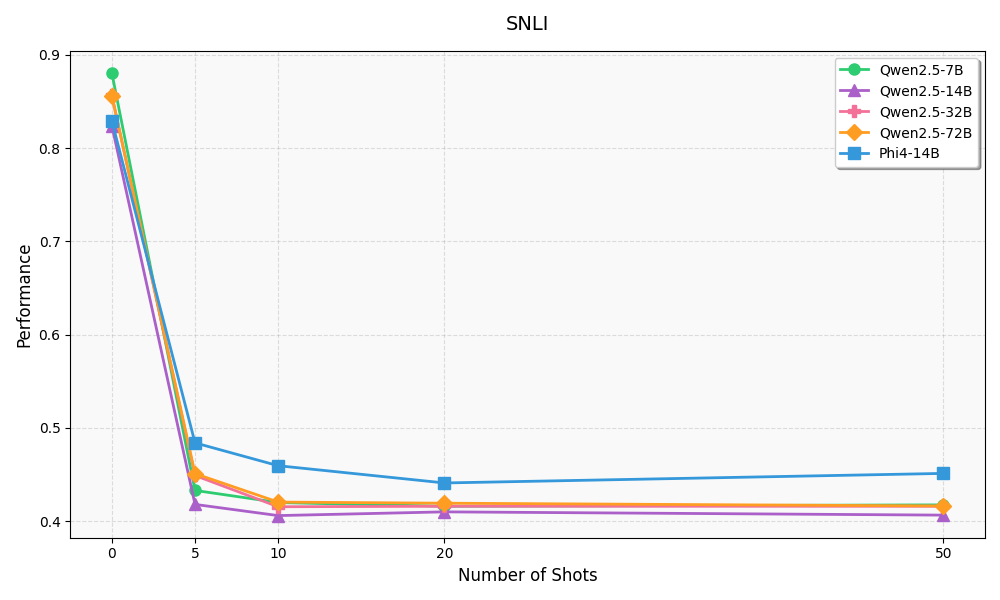}
\end{minipage}
}
\caption{Classification performance of models on the \data{SNLI} dataset under different few-shot learning scenarios ($n \in \{0,5,10,20,50\}$).}
\label{fig:snli_performance}
\end{figure*}

\begin{figure*}[h!]
    
    \centering

    \begin{tcolorbox}[colback=orange!20!white, colframe=black!20!orange, title=Prompt Instruction for Classification on AG News]
    You're given an input from the AG News dataset for article topic classification. You should classify it into one of the following categories: ``world'', ``sports'', ``business'', or ``science/technology''. Output the category only!
    \end{tcolorbox}

    \begin{tcolorbox}[colback=orange!20!white, colframe=black!20!orange, title=Prompt Instruction for Classification on SST2]
    You're given an input from the SST2 dataset for sentiment analysis. You should classify it into one of the following categories: ``positive'', or ``negative''. Output the category only!
    \end{tcolorbox}

    \begin{tcolorbox}[colback=orange!20!white, colframe=black!20!orange, title=Prompt Instruction for Classification on SNLI]
    You're given a premise and a hypothesis from the SNLI dataset for natural language inference. You should classify it into one of the following categories: ``neutral'', ``contradiction'', or ``entailment''. Output the category only!
    \end{tcolorbox}
    
    \caption{Prompt instruction for classification on \data{AG News}, \data{SST2}, and \data{SNLI}.
    }
    \label{fig:classification_prompt}
\end{figure*}

\section{Human Annotation}
\label{app:annotation}

Figure~\ref{fig:annotation_guidance} shows the annotation guideline provided to the recruited human annotators (\S\ref{subsec:user_study}). The counterfactuals are presented to annotators in the form of questionnaires. We use the Crowdee\footnote{\url{https://www.crowdee.com/}} crowdsourcing platform to recruit annotators, distribute the questionnaires, and store their responses. A total of 90 annotators were recruited, all of whom are native English speakers without requiring specific expertise in explainable AI (XAI). Each annotators will be given 15 counterfactuals, along with a set of predefined labels depending on datasets (\S\ref{subsec:datasets}). Each counterfactual will be evaluated by at least two annotators.

\begin{figure*}[h!]
    
    \centering

    \begin{tcolorbox}[colback=purple!20!white, colframe=purple!90!black, title=Annotation Guideline]

    \textbf{\#\#\# User Study Description}:
    
    Dear participants,
    
    Thanks for attending our user study. Our user study investigates how participants simulate model behavior based on provided explanations—an approach known as the simulatability test. You will be presented with explanations and predefined label options (depending on the dataset) and are asked to select the most appropriate label based solely on the explanations. We employ three datasets for this study: \textbf{AG News} (news topic classification), \textbf{SST-2} (sentiment analysis), and \textbf{SNLI} (natural language inference). The explanations are generated by models of varying sizes; however, model identities and sizes are not disclosed to participants.
    
    \bigskip
    
    \textbf{\#\#\# Dataset Structure:}

   \textbf{AG News:} This dataset consists of news articles. The task is to determine whether the topic of each article pertains to one of the following categories: Sports, World, Business, or Science/Technology.
   
    \bigskip
    \textbf{AG News Example}: \{example\}
    \bigskip

    \textbf{SST-2} (Stanford Sentiment Treebank): This dataset comprises movie reviews. The task is to assess the sentiment expressed in each review and classify it as either Positive or Negative.
    \bigskip
    
    \textbf{SST2 Example}: \{example\}
    \bigskip
    
    \textbf{SNLI} (Stanford Natural Language Inference): Each example consists of a premise and a hypothesis. The task is to determine the relationship between the two, categorizing it as either Entailment, Contradiction, or Neutral based on the information in the premise.
    \bigskip

    \textbf{SNLI Example}: \{example\}
    \bigskip
    
    \textit{Entailment} means the hypothesis must be true if the premise is true. \textit{Contradiction} means the hypothesis must be false if the premise is true. \textit{Neutral} means the hypothesis might be true, or might not — we can’t tell just from the premise.

    \bigskip
    
    \textbf{\#\#\# User Study Instruction:}
    
    You will be provided with 15 instances to evaluate. Each instance includes a single input field, depending on the dataset: a news article (AG News), a movie review (SST-2), or a premise–hypothesis pair (SNLI). If the text is applicable to multiple labels, please select the most appropriate one and report them in the follow-up question. If the text is applicable to multiple labels, please select the most appropriate one and report them in the follow-up question (``\textit{Do you think the text is applicable to multiple labels? (optional)}'').
    
    \end{tcolorbox}

    \caption{Annotation guidelines for the user study
    }
    \label{fig:annotation_guidance}
\end{figure*}

\section{Challenges in Label Flipping Identification}
\label{app:challenges}
Figure~\ref{fig:confusion} displays an example from \data{AG News}, its corresponding counterfactual generated by \lm{Llama3-70B} using \sys{FIZLE}, and the chain-of-thought from \lm{DeepSeek-R1-Distill-Llama-70B} \cite{deepseekai2025deepseekr1incentivizingreasoningcapability} which serves as the judge model $LLM_\mathcal{J}$ to identify the label flipping. Underlined words are determined by \lm{Llama3-70B} and newly inserted to achieve the necessary label flip. 

In the given example, \textbf{business}-related terms such as ``stock market'' and ``National Exchange'' are deliberately inserted in an attempt to flip the label from \textbf{sports} to \textbf{business}. However, the sentence still centers around a baseball game, prominently featuring ``Randy Johnson'', a well-known professional pitcher. This kind of example introduces a unique challenge. Human evaluators may recognize Randy Johnson as a sports figure and discount the inserted business terms. In contrast, LLMs may weigh both the artificial business cues and the surrounding sports context, attempting to reconcile the conflicting signals using their implicit knowledge. This divergence can lead to different forms of error:
\begin{itemize}
\item Human evaluators may rely more on surface-level keywords and could be misled by terms like ``stock market'', especially if they lack specific domain knowledge.

\item LLMs, on the other hand, may attempt to resolve the contradiction by grounding entities in factual knowledge, leading to confusion when contextual signals conflict.

\end{itemize}
Such discrepancies may explain observed patterns in the automatic evaluation (Table~\ref{tab:automatic_evaluation}). For instance, the 100\% label flip rate (LFR) by humans on  counterfactuals from the \data{SNLI} dataset suggests that annotators are highly influenced by inserted keywords. Meanwhile, LLMs exhibit lower FLRs, likely due to their more nuanced, knowledge-driven reasoning process. This illustrates a key distinction in how humans and models handle ambiguous inputs: humans may overfit to superficial cues, while LLMs attempt to resolve deeper semantic inconsistencies.

Furthermore, from the reasoning chains of \lm{DeepSeek-R1-Distill-Llama-70B} in Figure~\ref{fig:confusion}, we observe that it becomes confused when determining whether the label of the counterfactual flips, as the context remains ambiguous despite the inclusion of additional information about the stock market. This indicates that more advanced techniques for counterfactual generation are needed, and greater attention should be devoted to this area.

\begin{figure*}[h!]
    
    \centering

    \begin{tcolorbox}[colback=black!20!white, colframe=black!80!white, title=DeepSeek-R1-Distill-Llama-70B Evaluation]
     \textit{\textbf{Original input:}} Johnson Back to His Best as D-Backs End Streak  NEW YORK (Reuters) - Randy Johnson struck out 14 batters in  8 1/3 innings to help the Arizona Diamondbacks end a nine-game  losing streak with a 2-0 win over the host New York Mets in the  National League Sunday.

     \bigskip 
    \textbf{\textit{Label:}} \textbf{sports}
     \bigskip 
     
    \textbf{\textit{Counterfactual:}} Johnson Back to His Best as D-Backs End Streak \underline{in Stock Market}  NEW YORK (Reuters) - Randy Johnson struck out 14 \underline{deals} in  8 1/3 \underline{quarters} to help the Arizona Diamondbacks end a nine-game  losing streak \underline{in the stock market} with a 2-0 win over the host New York Mets in the National \underline{Exchange} Sunday.

    \bigskip
 
    \textbf{\textit{Chain of Thought}}: Alright, I'm trying to figure out how to classify this article into one of the four categories: world, sports, business, or science/technology. Let me read the input carefully.

    \bigskip

     Randy Johnson is a well-known baseball pitcher, so that's definitely a \textbf{sports} context. The content says "end a nine-game losing streak in the stock market". Wait, that's confusing because the stock market is related to business. Looking at the content, it says the Arizona Diamondbacks beating the New York Mets in the National Exchange Sunday. National Exchange could refer to a stock exchange. Therefore, the category should be \textbf{business}. 
    
    \end{tcolorbox}

    \caption{An example from \data{AG News}, its corresponding counterfactual generated by \sys{FIZLE}, and the chain-of-thought from \lm{DeepSeek-R1-Distill-Llama-70B}. \underline{Underlines} indicate the insertion of new words, compared to the original input.
    }
    \label{fig:confusion}
\end{figure*}

\section{Automatic Evaluation}
\label{apdx:auto_eval}
\subsection{Average Ranking}

\begin{table*}[t!]
    \centering
    \setlength{\extrarowheight}{2pt}
    \renewcommand*{\arraystretch}{0.65}
    
    \footnotesize
    
    \begin{minipage}{0.49\textwidth}
    \resizebox{\textwidth}{!}{%
        \begin{tabular}{cccccc}

        \toprule[1.5pt]
        Model & Relation. & \data{AG News}  & \data{SST2}  & \data{SNLI} & Avg. \\ 
        
        \hline

      \multirow{6}{*}{\rotatebox[origin=r]{90}{\tiny{\lm{Llama3-8B}}}} 
& \cellcolor[HTML]{9EC6F3}$\mathcal{R}_{sm}$ & 1 & 7 & 4 & \cellcolor[HTML]{aae5b5}4 \\
& \cellcolor[HTML]{E0E6FF}$\mathcal{R}_{dm}$ & 3 & 9 & 3 & 5\\
& \cellcolor[HTML]{FFF1D5}$\mathcal{R}_{sf}$ & 4 & 8 & 1 & 4.33\\
& \cellcolor[HTML]{febafb}$\mathcal{R}_{imw}$ & 8 & 3 & 8 & 6.33\\
& \cellcolor[HTML]{D8D8D8}$\mathcal{R}_{imwo}$ & 4.33 & 3.67 & 4.33 & 4.11\\
&  \cellcolor[HTML]{dfa8fe}\lm{Ensemble} & 8 & 4 & 8 & \cellcolor[HTML]{FEA8A8}6.67\\
\hline

\multirow{6}{*}{\rotatebox[origin=r]{90}{\tiny{\lm{Qwen2.5-14B}}}} 
& \cellcolor[HTML]{9EC6F3}$\mathcal{R}_{sm}$ & 4 & 8 & 2 & 4.67\\
& \cellcolor[HTML]{E0E6FF}$\mathcal{R}_{dm}$ & 3 & 9 & 3 & 5\\
& \cellcolor[HTML]{FFF1D5}$\mathcal{R}_{sf}$ & 2 & 6 & 4 & \cellcolor[HTML]{aae5b5}4 \\
& \cellcolor[HTML]{febafb}$\mathcal{R}_{imw}$ & 8 & 2 & 8 & \cellcolor[HTML]{FEA8A8}6\\
& \cellcolor[HTML]{D8D8D8}$\mathcal{R}_{imwo}$ & 4 & 5.33 & 4 & 4.44 \\
&  \cellcolor[HTML]{dfa8fe}\lm{Ensemble} & 8 & 2 & 8 & 6\\
\hline

\multirow{6}{*}{\rotatebox[origin=r]{90}{\tiny{\lm{Qwen2.5-32B}}}} 
& \cellcolor[HTML]{9EC6F3}$\mathcal{R}_{sm}$ & 6 & 7 & 6 & \cellcolor[HTML]{FEA8A8}6.33\\
& \cellcolor[HTML]{E0E6FF}$\mathcal{R}_{dm}$ & 1 & 8 & 5 & 4.67\\
& \cellcolor[HTML]{FFF1D5}$\mathcal{R}_{sf}$ & 7 & 6 & 4 & 5.67\\
& \cellcolor[HTML]{febafb}$\mathcal{R}_{imw}$ & 6.5 & 2.5 & 8.5 & 5.83\\
& \cellcolor[HTML]{D8D8D8}$\mathcal{R}_{imwo}$ & 3.33 & 4.67 & 4 & \cellcolor[HTML]{aae5b5}4\\
&  \cellcolor[HTML]{dfa8fe}\lm{Ensemble} & 8 & 5 & 1 & 4.67\\
\hline

\multirow{6}{*}{\rotatebox[origin=r]{90}{\tiny{\lm{Llama3-70B}}}} 
& \cellcolor[HTML]{9EC6F3}$\mathcal{R}_{sm}$ & 3 & 9 & 2& 4.67 \\
& \cellcolor[HTML]{E0E6FF}$\mathcal{R}_{dm}$ & 4 & 7 & 3 & 4.67\\
& \cellcolor[HTML]{FFF1D5}$\mathcal{R}_{sf}$ & 2 & 6 & 5 & 4.33\\
& \cellcolor[HTML]{febafb}$\mathcal{R}_{imw}$ & 8 & 4.5 & 8 & \cellcolor[HTML]{FEA8A8}6.83\\
& \cellcolor[HTML]{D8D8D8}$\mathcal{R}_{imwo}$ & 4 & 3.67 & 6 & 4.56\\
&  \cellcolor[HTML]{dfa8fe}\lm{Ensemble} & 8 & 3 & 1  & \cellcolor[HTML]{aae5b5}4\\

        \toprule[1.5pt]     
        
        \end{tabular}
        }
        \subcaption{Counterfactual examples generated using \sys{FIZLE}.}
    \end{minipage}
    \hspace{0.1cm}
    \begin{minipage}{0.49\textwidth}
    \resizebox{\textwidth}{!}{%
        \begin{tabular}{cccccc}

        \toprule[1.5pt]
        Model & Relation.  &  \data{AG News}  & \data{SST2}  & \data{SNLI} & Avg.\\ 
        
        \hline
                 \multirow{6}{*}{\rotatebox[origin=r]{90}{\tiny{\lm{Llama3-8B}}}} 
& \cellcolor[HTML]{9EC6F3}$\mathcal{R}_{sm}$ & 2 & 9 & 7 & 6 \\
& \cellcolor[HTML]{E0E6FF}$\mathcal{R}_{dm}$ & 3 & 1 & 8 & 4\\
& \cellcolor[HTML]{FFF1D5}$\mathcal{R}_{sf}$ & 6 & 3 & 3 & \cellcolor[HTML]{aae5b5}4\\
& \cellcolor[HTML]{febafb}$\mathcal{R}_{imw}$ & 8 & 5.5 & 5 & \cellcolor[HTML]{FEA8A8}6.17\\
& \cellcolor[HTML]{D8D8D8}$\mathcal{R}_{imwo}$ & 3.33 & 5.67 & 5.33 & 4.78\\
&  \cellcolor[HTML]{dfa8fe}\lm{Ensemble} & 8 & 4 & 1 & 4.33\\
\hline

\multirow{6}{*}{\rotatebox[origin=r]{90}{\tiny{\lm{Qwen2.5-14B}}}} 
& \cellcolor[HTML]{9EC6F3}$\mathcal{R}_{sm}$ & 6 & 8 & 4 & 6\\
& \cellcolor[HTML]{E0E6FF}$\mathcal{R}_{dm}$ & 5 & 9 & 8 & \cellcolor[HTML]{FEA8A8}7.33\\
& \cellcolor[HTML]{FFF1D5}$\mathcal{R}_{sf}$ & 4 & 7 & 7 & 6\\
& \cellcolor[HTML]{febafb}$\mathcal{R}_{imw}$ & 8 & 3.5 & 3.5 & 5\\
& \cellcolor[HTML]{D8D8D8}$\mathcal{R}_{imwo}$ & 2 & 3.33 & 5.33 & \cellcolor[HTML]{aae5b5}3.55\\
&  \cellcolor[HTML]{dfa8fe}\lm{Ensemble} & 8 & 4 & 3 & 5\\
\hline

\multirow{6}{*}{\rotatebox[origin=r]{90}{\tiny{\lm{Qwen2.5-32B}}}} 
& \cellcolor[HTML]{9EC6F3}$\mathcal{R}_{sm}$ & 5 & 6 & 4 & 5\\
& \cellcolor[HTML]{E0E6FF}$\mathcal{R}_{dm}$ & 4 & 9 & 8 & \cellcolor[HTML]{FEA8A8}7\\
& \cellcolor[HTML]{FFF1D5}$\mathcal{R}_{sf}$ & 2 & 8 & 5 & 5\\
& \cellcolor[HTML]{febafb}$\mathcal{R}_{imw}$ & 8.5 & 2.5 & 4.5 & 5.17\\
& \cellcolor[HTML]{D8D8D8}$\mathcal{R}_{imwo}$ & 3.33 & 4.33 & 4 & \cellcolor[HTML]{aae5b5}3.89\\
&  \cellcolor[HTML]{dfa8fe}\lm{Ensemble} & 7 & 4 & 7 & 6\\
\hline

\multirow{6}{*}{\rotatebox[origin=r]{90}{\tiny{\lm{Llama3-70B}}}} 
& \cellcolor[HTML]{9EC6F3}$\mathcal{R}_{sm}$ & 6 & 8 & 4 & 6\\
& \cellcolor[HTML]{E0E6FF}$\mathcal{R}_{dm}$ & 4 & 9 & 8 & \cellcolor[HTML]{FEA8A8}7\\
& \cellcolor[HTML]{FFF1D5}$\mathcal{R}_{sf}$ & 2 & 2 & 3 & \cellcolor[HTML]{aae5b5}2.33\\
& \cellcolor[HTML]{febafb}$\mathcal{R}_{imw}$ & 8 & 6 & 3 & 5.67\\
& \cellcolor[HTML]{D8D8D8}$\mathcal{R}_{imwo}$ & 3 & 2.67 & 6 & 3.89\\
&  \cellcolor[HTML]{dfa8fe}\lm{Ensemble} & 8 & 6 & 6 & 6.67\\

        \toprule[1.5pt]     
        
        \end{tabular}
        }
        \subcaption{Counterfactual examples generated using FLARE.}
    \end{minipage}
    \caption{The average ranking of judge-generator model relationship based on $\Delta$ in Table~\ref{tab:automatic_evaluation_values} (lower rankings indicate better alignment). Counterfactual examples are generated by \lm{Qwen2.5-\{14B,32B\}} and \lm{Llama3-\{8B,70B\}}, evaluated across judge models exhibiting \colorbox{lightblue}{same}, \colorbox{blue}{distilled}, \colorbox{lighpurple}{same family}, and independent \colorbox{purple}{w/} and \colorbox{gray}{w/o} fine-tuning relationships on \data{AG News}, \data{SST2} and \data{SNLI}. \colorbox{deepgreen}{Green-highlighted} values indicate that the judge model with the given relationship aligns \textbf{closely} with human annotators, while \colorbox{deepred}{red-highlighted} values indicate the \textbf{opposite}.
    }
    \label{tab:automatic_evaluation}
\end{table*}
Table~\ref{tab:automatic_evaluation} shows the average ranking of each judge-generator model relationship based on $\Delta$ in Table~\ref{tab:automatic_evaluation_values}. Figure~\ref{fig:figure} shows the average ranking and $r_{m/\ell}$ of the relationships $\mathcal{R}_{\mathcal{G}, \mathcal{J}}$. We observe that judge models with $\mathcal{R}_{imwo}$ achieve the lowest average ranking, while those with $\mathcal{R}_{imw}$ achieve the highest. Here, a lower ranking indicates that the corresponding label-flipping results are more closely aligned with human evaluation outcomes.

\subsection{LFR Differences in Discrete Values}
\begin{table*}[t!]
    \centering
    \setlength{\extrarowheight}{2pt}
    \renewcommand*{\arraystretch}{1}
    
    \footnotesize
    
    \begin{minipage}{0.49\textwidth}
    \resizebox{\textwidth}{!}{%
        \begin{tabular}{cp{1cm}cccc}

        \toprule[1.5pt]
        Model & Relation. & Judge Model ($LLM_\mathcal{J}$) &  \data{AG News}  & \data{SST2}  & \data{SNLI} \\ 
        
        \hline

       
      \centering \multirow{10}{*}{\rotatebox[origin=r]{90}{\lm{Llama3-8B}}} 
& $\mathcal{R}_{sm}$ & \cellcolor[HTML]{9EC6F3}\lm{Llama3-8B} & \cellcolor[HTML]{FF8C00}$+15.36$  & $-23.53$  & $+27.00$ \\

& \textbf{$\mathcal{R}_{dm}$} & \cellcolor[HTML]{E0E6FF}\tiny{\lm{DeepSeek-R1-Distill-Llama-8B}} & $+17.36$  & \cellcolor[HTML]{C1AFEC}$-27.13$  & $+26.80$ \\

& $\mathcal{R}_{sf}$  & \cellcolor[HTML]{FFF1D5}\lm{Llama3-70B} & $+22.56$  & $-25.53$  & \cellcolor[HTML]{FF8C00}$+20.80$ \\

& $\mathcal{R}_{imw}$ & \cellcolor[HTML]{febafb}\lm{BERT} & $+36.56$  & \cellcolor[HTML]{FF8C00}$-15.33$  & $+32.80$ \\  

& $\mathcal{R}_{imw}$ & \cellcolor[HTML]{febafb}\lm{RoBERTa} & \cellcolor[HTML]{C1AFEC}$+37.56$  & $-23.13$  & \cellcolor[HTML]{C1AFEC}$+35.80$ \\


& $\mathcal{R}_{imwo}$& \cellcolor[HTML]{d8d8d8}\lm{Phi4-14B} & $+15.56$  & $-23.53$  & $+26.80$ \\  

& $\mathcal{R}_{imwo}$ & \cellcolor[HTML]{d8d8d8}\lm{Mistral-Large} & $+23.56$  & $-21.13$  & $+32.40$ \\  

& $\mathcal{R}_{imwo}$ & \cellcolor[HTML]{d8d8d8}\lm{Gemini-1.5-pro}& $+28.32$  & $-17.65$  & $+30.60$  \\

& - & \cellcolor[HTML]{dfa8fe}\lm{Ensemble} & $+37.56$ & $-23.13$  & $+35.80$ \\

& - & \cellcolor[HTML]{F2CFC2}\lm{User Study}& \cellcolor[HTML]{F2CFC2}$75.56$  & \cellcolor[HTML]{F2CFC2}$46.67$  & \cellcolor[HTML]{F2CFC2}$100.00$  \\

  \hline
       
    \centering \multirow{10}{*}{\rotatebox[origin=r]{90}{\lm{Qwen2.5-14B}}} 
& $\mathcal{R}_{sm}$ & \cellcolor[HTML]{9EC6F3}\lm{Qwen2.5-14B}     & $+21.24$  & $-28.62$  & $+17.80$ \\
        
& $\mathcal{R}_{dm}$ & \cellcolor[HTML]{E0E6FF}\tiny{\lm{DeepSeek-R1-Distill-Qwen-14B}} & $+20.64$  &\cellcolor[HTML]{C1AFEC} $-29.02$  & $+18.40$ \\

& $\mathcal{R}_{sf}$& \cellcolor[HTML]{FFF1D5}\lm{Qwen2.5-72B} & $+17.04$  & $-28.02$  & $+21.20$ \\
        
& $\mathcal{R}_{imw}$& \cellcolor[HTML]{febafb}\lm{BERT} & \cellcolor[HTML]{C1AFEC}$+46.04$  & \cellcolor[HTML]{FF8C00}$-23.02$  & \cellcolor[HTML]{C1AFEC}$+31.80$  \\  

& $\mathcal{R}_{imw}$& \cellcolor[HTML]{febafb}\lm{RoBERTa} & $+43.84$  & $-24.02$  & $+38.60$ \\


& $\mathcal{R}_{imwo}$& \cellcolor[HTML]{d8d8d8}\lm{Phi4-14B}& \cellcolor[HTML]{FF8C00}$+14.84$  & $-27.62$  & \cellcolor[HTML]{FF8C00}$+13.80$  \\

& $\mathcal{R}_{imwo}$ & \cellcolor[HTML]{d8d8d8}\lm{Mistral-Large}& $+27.04$  & $-28.42$  & $+22.00$  \\

& $\mathcal{R}_{imwo}$ & \cellcolor[HTML]{d8d8d8}\lm{Gemini-1.5-pro}& $+22.13$  & $-26.62$  & $+27.20$  \\

& - & \cellcolor[HTML]{dfa8fe}\lm{Ensemble} & $+43.84$ & $-24.02$ & $+32.02$ \\

& - & \cellcolor[HTML]{F2CFC2}\lm{User Study}& \cellcolor[HTML]{F2CFC2}$84.44$  & \cellcolor[HTML]{F2CFC2}$57.78$  & \cellcolor[HTML]{F2CFC2}$100.00$  \\
       
       \hline

       \centering 
\multirow{11}{*}{\rotatebox[origin=r]{90}{\lm{Qwen2.5-32B}}} 
& $\mathcal{R}_{sm}$ & \cellcolor[HTML]{9EC6F3}\lm{Qwen2.5-32B}     & $-10.33$ & $-22.13$  & $+30.00$  \\
        
& $\mathcal{R}_{dm}$& \cellcolor[HTML]{E0E6FF}\tiny{\lm{DeepSeek-R1-Distill-Qwen-32B}} & \cellcolor[HTML]{FF8C00}$-0.53$  & $-22.33$  & $+25.60$ \\

& $\mathcal{R}_{sf}$ & \cellcolor[HTML]{FFF1D5}\lm{Qwen2.5-72B} & $-16.21$  & $-21.73$  & $+23.80$ \\
        
& $\mathcal{R}_{imw}$& \cellcolor[HTML]{febafb}\lm{BERT} & $+9.62$  & $-19.53$  & $+33.00$ \\  

& $\mathcal{R}_{imw}$& \cellcolor[HTML]{febafb}\lm{RoBERTa} & \cellcolor[HTML]{C1AFEC}$+27.62$  & \cellcolor[HTML]{FF8C00}$-15.13$  & \cellcolor[HTML]{C1AFEC}$+38.20$ \\


& $\mathcal{R}_{imwo}$ & \cellcolor[HTML]{d8d8d8}\lm{Phi4-14B} & $-3.98$  & \cellcolor[HTML]{C1AFEC}$-22.93$  & $+15.80$ \\  

& $\mathcal{R}_{imwo}$ & \cellcolor[HTML]{d8d8d8}\lm{Mistral-Large} & $+10.22$  & $-18.13$  & $+22.60$ \\  

& $\mathcal{R}_{imwo}$ & \cellcolor[HTML]{d8d8d8}\lm{Gemini-1.5-pro}& $+3.40$  & $-18.76$  & $+30.60$  \\

& - & \cellcolor[HTML]{dfa8fe}\lm{Ensemble} & $+27.62$ & $-20.93$ & \cellcolor[HTML]{FF8C00}$+12.80$\\

& - & \cellcolor[HTML]{F2CFC2}\lm{User Study}& \cellcolor[HTML]{F2CFC2}$62.22$  & \cellcolor[HTML]{F2CFC2}$66.67$  & \cellcolor[HTML]{F2CFC2}$100.00$  \\

       \hline

      \centering \multirow{10}{*}{\rotatebox[origin=r]{90}{\lm{Llama3-70B}}}& $\mathcal{R}_{sm}$ & \cellcolor[HTML]{9EC6F3}\lm{Llama3-70B} & $+1.84$  & \cellcolor[HTML]{C1AFEC}$-43.38$  & $+16.40$ \\
        
        & $\mathcal{R}_{dm}$ & \cellcolor[HTML]{E0E6FF}\tiny{\lm{DeepSeek-R1-Distill-Llama-70B}} & $+3.84$  & $-41.70$  & $+18.40$ \\

         & $\mathcal{R}_{sf}$& \cellcolor[HTML]{FFF1D5}\lm{Llama3-8B} & $-1.76$  & $-40.58$  & $+26.20$ \\
        
       & $\mathcal{R}_{imw}$& \cellcolor[HTML]{febafb}\lm{BERT}& $+18.24$  & $-36.98$  & $+31.40$  \\
      & $\mathcal{R}_{imw}$ & \cellcolor[HTML]{febafb}\lm{RoBERTa} & \cellcolor[HTML]{C1AFEC}$+21.24$  & $-36.78$  & \cellcolor[HTML]{C1AFEC}$+37.40$ \\ 

      
      & $\mathcal{R}_{imwo}$ & \cellcolor[HTML]{d8d8d8}\lm{Phi4-14B} & \cellcolor[HTML]{FF8C00}$+0.84$  & $-42.38$  & $+22.80$ \\  
      
       & $\mathcal{R}_{imwo}$ & \cellcolor[HTML]{d8d8d8}\lm{Mistral-Large} & $+11.64$  & \cellcolor[HTML]{FF8C00}$-31.38$  & $+29.80$ \\  
       
        & $\mathcal{R}_{imwo}$ & \cellcolor[HTML]{d8d8d8}\lm{Gemini-1.5-pro}& $-10.67$  & $-35.17$  & $+31.40$  \\

        & - & \cellcolor[HTML]{dfa8fe}\lm{Ensemble} & $+21.24$ & $-36.78$ & \cellcolor[HTML]{FF8C00}$+9.80$ \\

        & - & \cellcolor[HTML]{F2CFC2}\lm{User Study}& \cellcolor[HTML]{F2CFC2}$64.44$  & \cellcolor[HTML]{F2CFC2}$42.22$  & \cellcolor[HTML]{F2CFC2}$100.00$  \\

        \toprule[1.5pt]     
        
        \end{tabular}
        }
        \subcaption{Counterfactual examples generated using \sys{FIZLE}.}
    \end{minipage}
    \hspace{0.1cm}
    \begin{minipage}{0.49\textwidth}
    \resizebox{\textwidth}{!}{%
        \begin{tabular}{cp{1cm}cccc}

        \toprule[1.5pt]
        Model & Relation. & Judge Model ($LLM_\mathcal{J}$) &  \data{AG News}  & \data{SST2}  & \data{SNLI} \\ 
        
        \hline
    
        

        
       
      
       


       
       \centering \multirow{10}{*}{\rotatebox[origin=r]{90}{\lm{Llama3-8B}}} 
& $\mathcal{R}_{sm}$ & \cellcolor[HTML]{9EC6F3}\lm{Llama3-8B} & $+50.27$ & \cellcolor[HTML]{C1AFEC}$11.29$ & $+34.86$ \\

& $\mathcal{R}_{dm}$ & \cellcolor[HTML]{E0E6FF}\tiny{\lm{DeepSeek-R1-Distill-Llama-8B}} & $+54.87$ & \cellcolor[HTML]{FF8C00}$-1.10$ & \cellcolor[HTML]{C1AFEC}$+40.46$ \\

& $\mathcal{R}_{sf}$ & \cellcolor[HTML]{FFF1D5}\lm{Llama3-70B} & $+64.47$ & $+4.29$ & $+29.26$ \\
& $\mathcal{R}_{imw}$ & \cellcolor[HTML]{febafb}\lm{BERT} & \cellcolor[HTML]{C1AFEC}$+68.27$ & $+8.77$ & $+31.04$ \\
& $\mathcal{R}_{imw}$ & \cellcolor[HTML]{febafb}\lm{RoBERTa} & $+68.07$ & $+6.24$ & $+29.77$ \\


& $\mathcal{R}_{imwo}$ & \cellcolor[HTML]{d8d8d8}\lm{Phi4-14B} & $+56.07$ & $+3.61$ & $+28.50$ \\
& $\mathcal{R}_{imwo}$ & \cellcolor[HTML]{d8d8d8}\lm{Mistral-Large} & $+58.47$ & $+9.34$ & $+62.34$ \\
& $\mathcal{R}_{imwo}$ & \cellcolor[HTML]{d8d8d8}\lm{Gemini-1.5-pro}& \cellcolor[HTML]{FF8C00}$+39.43$  & $+9.01$  & \cellcolor[HTML]{FF8C00}$+30.60$  \\

& - & \cellcolor[HTML]{dfa8fe}\lm{Ensemble} & $+68.07$ & $+6.24$ & $+14.25$ \\

& - & \cellcolor[HTML]{F2CFC2}\lm{User Study} & \cellcolor[HTML]{F2CFC2}$86.67$ & \cellcolor[HTML]{F2CFC2}$73.33$ & \cellcolor[HTML]{F2CFC2}$100.00$ \\
\hline

    \centering \multirow{10}{*}{\rotatebox[origin=r]{90}{\lm{Qwen2.5-14B}}}& $\mathcal{R}_{sm}$ & \cellcolor[HTML]{9EC6F3}\lm{Qwen2.5-14B}     & $46.93$  & $-22.44$  & $+32.29$ \\
        
        & $\mathcal{R}_{dm}$ & \cellcolor[HTML]{E0E6FF}\tiny{\lm{DeepSeek-R1-Distill-Qwen-14B}} & $+42.13$  & \cellcolor[HTML]{C1AFEC}$-24.16$  & $+46.71$ \\

         & $\mathcal{R}_{sf}$& \cellcolor[HTML]{FFF1D5}\lm{Qwen2.5-72B} & $+41.93$  & $-22.44$  & $+36.05$ \\
        
       & $\mathcal{R}_{imw}$ & \cellcolor[HTML]{febafb}\lm{BERT} & \cellcolor[HTML]{C1AFEC}$+53.53$  & $-16.93$  & $+30.09$  \\ 
       
       & $\mathcal{R}_{imw}$ & \cellcolor[HTML]{febafb}\lm{RoBERTa} & $+53.13$  & $-19.00$  & $+33.23$ \\

       
       & $\mathcal{R}_{imwo}$ & \cellcolor[HTML]{d8d8d8}\lm{Phi4-14B}& $+41.13$  & $-22.44$  & $+33.23$  \\
       
       & $\mathcal{R}_{imwo}$ & \cellcolor[HTML]{d8d8d8}\lm{Mistral-Large}& $+40.73$  & \cellcolor[HTML]{FF8C00}$-12.46$  & \cellcolor[HTML]{C1AFEC}$+54.55$  \\
       
        & $\mathcal{R}_{imwo}$ & \cellcolor[HTML]{d8d8d8}\lm{Gemini-1.5-pro}& \cellcolor[HTML]{FF8C00}$+11.02$  & $-17.75$  & \cellcolor[HTML]{FF8C00}$+27.20$  \\

        & - & \cellcolor[HTML]{dfa8fe}\lm{Ensemble} & $+53.13$ & $-19.00$ & $+33.23$ \\

       & - & \cellcolor[HTML]{F2CFC2}\lm{User Study}& \cellcolor[HTML]{F2CFC2}$73.33$  & \cellcolor[HTML]{F2CFC2}$66.67$  & \cellcolor[HTML]{F2CFC2}$100.00$  \\

       \hline

       \centering \multirow{10}{*}{\rotatebox[origin=r]{90}{\lm{Qwen2.5-32B}}} 
& $\mathcal{R}_{sm}$ & \cellcolor[HTML]{9EC6F3}\lm{Qwen2.5-32B} & $+38.31$ & $-35.36$ & $+32.54$ \\

& $\mathcal{R}_{dm}$ & \cellcolor[HTML]{E0E6FF}\tiny{\lm{DeepSeek-R1-Distill-Qwen-32B}} & $+37.71$ & \cellcolor[HTML]{C1AFEC}$-39.49$ & $+50.36$ \\

& $\mathcal{R}_{sf}$ & \cellcolor[HTML]{FFF1D5}\lm{Qwen2.5-72B} & $+35.91$ & $-37.54$ & $+34.68$ \\

& $\mathcal{R}_{imw}$ & \cellcolor[HTML]{febafb}\lm{BERT} & \cellcolor[HTML]{C1AFEC}$+47.91$ & $-31.92$ & $+30.64$ \\

& $\mathcal{R}_{imw}$ & \cellcolor[HTML]{febafb}\lm{RoBERTa} & $+47.41$ & $-32.26$ & $+34.92$ \\


& $\mathcal{R}_{imwo}$ & \cellcolor[HTML]{d8d8d8}\lm{Phi4-14B} & $+36.71$ & $-36.39$ & $+30.64$ \\

& $\mathcal{R}_{imwo}$ & \cellcolor[HTML]{d8d8d8}\lm{Mistral-Large} & $+38.91$ & \cellcolor[HTML]{FF8C00}$-24.92$ & \cellcolor[HTML]{C1AFEC}$+53.21$ \\

& $\mathcal{R}_{imwo}$ & \cellcolor[HTML]{d8d8d8}\lm{Gemini-1.5-pro}& \cellcolor[HTML]{FF8C00}$+12.29$  & $-34.32$  & \cellcolor[HTML]{FF8C00}$+30.60$  \\

& - & \cellcolor[HTML]{dfa8fe}\lm{Ensemble} & $+47.31$ & $-32.26$ & $-34.92$\\

& - & \cellcolor[HTML]{F2CFC2}\lm{User Study} & \cellcolor[HTML]{F2CFC2}$71.11$ & \cellcolor[HTML]{F2CFC2}$51.11$ & \cellcolor[HTML]{F2CFC2}$100.00$ \\
\hline

       \centering \multirow{10}{*}{\rotatebox[origin=r]{90}{\lm{Llama3-70B}}} 
& $\mathcal{R}_{sm}$ & \cellcolor[HTML]{9EC6F3}\lm{Llama3-70B} & $+39.13$ & $-25.05$ & $+36.52$ \\

& $\mathcal{R}_{dm}$ & \cellcolor[HTML]{E0E6FF}\tiny{\lm{DeepSeek-R1-Distill-Llama-70B}} & $+34.73$ & \cellcolor[HTML]{C1AFEC}$-27.11$ & $+44.58$ \\

& $\mathcal{R}_{sf}$ & \cellcolor[HTML]{FFF1D5}\lm{Llama3-8B} & $+26.53$ & $-12.09$ & $+33.25$ \\

& $\mathcal{R}_{imw}$  & \cellcolor[HTML]{febafb}\lm{BERT} & \cellcolor[HTML]{C1AFEC}$+42.13$ & $-22.53$ & \cellcolor[HTML]{FF8C00}$+30.48$ \\

& $\mathcal{R}_{imw}$ & \cellcolor[HTML]{febafb}\lm{RoBERTa} & $+39.73$ & $-23.67$ & $+37.53$ \\


& $\mathcal{R}_{imwo}$ & \cellcolor[HTML]{d8d8d8}\lm{Phi4-14B} & $+33.53$ & $-20.69$ & $+37.78$ \\

& $\mathcal{R}_{imwo}$ & \cellcolor[HTML]{d8d8d8}\lm{Mistral-Large} & $+37.53$ & \cellcolor[HTML]{FF8C00}$-11.63$ & \cellcolor[HTML]{C1AFEC}$+47.60$ \\

& $\mathcal{R}_{imwo}$ & \cellcolor[HTML]{d8d8d8}\lm{Gemini-1.5-pro}& \cellcolor[HTML]{FF8C00}$+19.56$  & $-15.17$  & $+31.40$  \\

& - & \cellcolor[HTML]{dfa8fe}\lm{Ensemble} & $+39.73$ & $-23.67$ & $+37.53$ \\

& - & \cellcolor[HTML]{F2CFC2}\lm{User Study} & \cellcolor[HTML]{F2CFC2}$73.33$ & \cellcolor[HTML]{F2CFC2}$62.22$ & \cellcolor[HTML]{F2CFC2}$100.00$ \\
\hline

        \toprule[1.5pt]     
        
        \end{tabular}
        }
        \subcaption{Counterfactual examples generated using FLARE.}
    \end{minipage}
    \caption{The LFR difference ($\Delta\%$) between the \colorbox{lightred}{user study} and the judge–generated model relationships (with values closer to 0 indicating better alignment). Counterfactual examples are generated by \lm{Qwen2.5-\{14B,32B\}} and \lm{Llama3-\{8B,70B\}}, evaluated across judge models exhibiting \colorbox{lightblue}{same}, \colorbox{blue}{distilled}, \colorbox{lighpurple}{same family}, and independent \colorbox{purple}{w/} and \colorbox{gray}{w/o} fine-tuning relationships on \data{AG News}, \data{SST2} and \data{SNLI}. The \colorbox{lightred}{user study} to assess the LFR is conducted on 45 selected counterfactuals (\S\ref{subsec:user_study}).  \colorbox{deepred}{Red-highlighted} values indicate that the judge model with the given relationship aligns \textbf{closely} with human annotators, while \colorbox{deepgreen}{green-highlighted} values indicate the \textbf{opposite}.
    }
    \label{tab:automatic_evaluation_values}
\end{table*}
\label{app:automatic_evaluation_values}

Table~\ref{tab:automatic_evaluation_values} shows the differences calculated by subtracting the user study results from the results of the judge models $LLM_\mathcal{J}$ results.

\subsection{A Sanity Check}
\label{app:sanity_check}
\begin{table*}[t!]
    \centering
    \setlength{\extrarowheight}{2pt}
    \renewcommand*{\arraystretch}{1}
    
    \footnotesize
    
    \begin{minipage}{0.49\textwidth}
    \resizebox{\textwidth}{!}{%
        \begin{tabular}{ccccc}

        \toprule[1.5pt]
        Model & Judge Model ($LLM_\mathcal{J}$) &  \data{AG News}  & \data{SST2}  & \data{SNLI} \\ 
        
        \hline
    
        

        


       
      \centering \multirow{8}{*}{\rotatebox[origin=r]{90}{\lm{Llama3-8B}}} & \cellcolor[HTML]{9EC6F3}\lm{Llama3-8B} & \cellcolor[HTML]{FF8C00}$60.20$  & $70.20$  & $75.55$ \\
        
        & \cellcolor[HTML]{E0E6FF}\tiny{\lm{DeepSeek-R1-Distill-Llama-8B}} & $58.20$  & \cellcolor[HTML]{C1AFEC}$73.80$  & $71.11$ \\

         & \cellcolor[HTML]{FFF1D5}\lm{Llama3-70B} & $53.00$  & $72.20$  & \cellcolor[HTML]{FF8C00}$82.22$ \\
        
       & \cellcolor[HTML]{febafb}\lm{BERT} & $39.00$  & \cellcolor[HTML]{FF8C00}$62.00$  & $71.11$ \\  
      & \cellcolor[HTML]{febafb}\lm{RoBERTa} & \cellcolor[HTML]{C1AFEC}$38.00$  & $69.80$  & \cellcolor[HTML]{C1AFEC}$62.22$ \\
       & \cellcolor[HTML]{d8d8d8}\lm{Phi4-14B} & $60.00$  & $70.20$  & $68.88$ \\  
       & \cellcolor[HTML]{d8d8d8}\lm{Mistral-Large} & $52.00$  & $67.80$  & $62.22$ \\  
       
       & \cellcolor[HTML]{F2CFC2}\lm{User Study}& \cellcolor[HTML]{F2CFC2}$75.56$  & \cellcolor[HTML]{F2CFC2}$46.67$  & \cellcolor[HTML]{F2CFC2}$100.00$  \\
       
       \hline

    \centering \multirow{8}{*}{\rotatebox[origin=r]{90}{\lm{Qwen2.5-14B}}} & \cellcolor[HTML]{9EC6F3}\lm{Qwen2.5-14B}     & $57.77$  & $93.33$  & $84.44$ \\
        
        & \cellcolor[HTML]{E0E6FF}\tiny{\lm{DeepSeek-R1-Distill-Qwen-14B}} & $63.80$  & $95.55$  & \cellcolor[HTML]{FF8C00}$91.11$ \\

         & \cellcolor[HTML]{FFF1D5}\lm{Qwen2.5-72B} & $68.88$  & \cellcolor[HTML]{C1AFEC}$97.77$  & $80.00$ \\
        
       & \cellcolor[HTML]{febafb}\lm{BERT} & \cellcolor[HTML]{C1AFEC}$26.66$  & \cellcolor[HTML]{FF8C00}$86.66$  & $73.33$  \\  
      & \cellcolor[HTML]{febafb}\lm{RoBERTa} & $26.66$  & $86.66$  & \cellcolor[HTML]{C1AFEC}$57.77$ \\
       & \cellcolor[HTML]{d8d8d8}\lm{Phi4-14B}& \cellcolor[HTML]{FF8C00}$71.11$  & $93.33$  & $86.66$  \\
       & \cellcolor[HTML]{d8d8d8}\lm{Mistral-Large}& $62.22$  & $93.33$  & $82.22$  \\
       
       & \cellcolor[HTML]{F2CFC2}\lm{User Study}& \cellcolor[HTML]{F2CFC2}$84.44$  & \cellcolor[HTML]{F2CFC2}$57.78$  & \cellcolor[HTML]{F2CFC2}$100.00$  \\
       
       \hline

       \centering \multirow{8}{*}{\rotatebox[origin=r]{90}{\lm{Qwen2.5-32B}}} & \cellcolor[HTML]{9EC6F3}\lm{Qwen2.5-32B}     & $60.00$ & $91.11$  & $71.11$  \\
        
        & \cellcolor[HTML]{E0E6FF}\tiny{\lm{DeepSeek-R1-Distill-Qwen-32B}} & $62.75$  & $88.88$  & $71.11$ \\

         & \cellcolor[HTML]{FFF1D5}\lm{Qwen2.5-72B} & $62.22$  & \cellcolor[HTML]{C1AFEC}$91.11$  & $77.77$ \\
        
       & \cellcolor[HTML]{febafb}\lm{BERT} & \cellcolor[HTML]{C1AFEC}$31.11$  & $84.44$  & $71.11$ \\  
      & \cellcolor[HTML]{febafb}\lm{RoBERTa} & $33.33$  & \cellcolor[HTML]{FF8C00}$84.44$  & \cellcolor[HTML]{C1AFEC}$57.77$ \\
       & \cellcolor[HTML]{d8d8d8}\lm{Phi4-14B} & $71.11$  & $88.88$  & $82.22$ \\  
       & \cellcolor[HTML]{d8d8d8}\lm{Mistral-Large} & \cellcolor[HTML]{FF8C00}$66.66$  & $86.66$  & \cellcolor[HTML]{FF8C00}$84.44$ \\  
       
       & \cellcolor[HTML]{F2CFC2}\lm{User Study}& \cellcolor[HTML]{F2CFC2}$62.22$  & \cellcolor[HTML]{F2CFC2}$66.67$  & \cellcolor[HTML]{F2CFC2}$100.00$  \\
       \hline

       \centering \multirow{8}{*}{\rotatebox[origin=r]{90}{\lm{Llama3-70B}}} & \cellcolor[HTML]{9EC6F3}\lm{Llama3-70B} & $62.60$  & \cellcolor[HTML]{C1AFEC}$85.60$  & $84.44$ \\
        
        & \cellcolor[HTML]{E0E6FF}\tiny{\lm{DeepSeek-R1-Distill-Llama-70B}} & $60.60$  & $83.92$  & \cellcolor[HTML]{FF8C00}$86.66$ \\

         & \cellcolor[HTML]{FFF1D5}\lm{Llama3-8B} & $66.20$  & $82.80$  & $71.11$ \\
        
       & \cellcolor[HTML]{febafb}\lm{BERT}& $46.20$  & $79.20$  & $73.33$  \\
      & \cellcolor[HTML]{febafb}\lm{RoBERTa} & \cellcolor[HTML]{C1AFEC}$43.20$  & $79.00$  & \cellcolor[HTML]{C1AFEC}$62.60$ \\ 
      & \cellcolor[HTML]{d8d8d8}\lm{Phi4-14B} & \cellcolor[HTML]{FF8C00}$63.60$  & $84.60$  & $73.33$ \\  
       & \cellcolor[HTML]{d8d8d8}\lm{Mistral-Large} & $52.80$  & \cellcolor[HTML]{FF8C00}$73.60$  & $75.55$ \\  

        & \cellcolor[HTML]{F2CFC2}\lm{User Study}& \cellcolor[HTML]{F2CFC2}$64.44$  & \cellcolor[HTML]{F2CFC2}$42.22$  & \cellcolor[HTML]{F2CFC2}$100.00$  \\
       
        \toprule[1.5pt]     
        
        \end{tabular}
        }
        \subcaption{Counterfactual examples generated using \sys{FIZLE}.}
    \end{minipage}
    \hspace{0.1cm}
    \begin{minipage}{0.49\textwidth}
    \resizebox{\textwidth}{!}{%
        \begin{tabular}{ccccc}

        \toprule[1.5pt]
        Model & Judge Model ($LLM_\mathcal{J}$) &  \data{AG News}  & \data{SST2}  & \data{SNLI} \\ 
        
        \hline
    
        

        


       
       \centering \multirow{8}{*}{\rotatebox[origin=r]{90}{\lm{Llama3-8B}}} & \cellcolor[HTML]{9EC6F3}\lm{Llama3-8B} & \cellcolor[HTML]{FF8C00}$42.22$  & $80.00$  & $64.44$ \\
        
        & \cellcolor[HTML]{E0E6FF}\tiny{\lm{DeepSeek-R1-Distill-Llama-8B}} & $40.00$  & \cellcolor[HTML]{C1AFEC}$82.22$  & \cellcolor[HTML]{C1AFEC}$55.55$ \\

         & \cellcolor[HTML]{FFF1D5}\lm{Llama3-70B} & $33.33$  & $73.33$  & $73.33$ \\
        
       & \cellcolor[HTML]{febafb}\lm{BERT} & \cellcolor[HTML]{C1AFEC}$15.55$  & $71.11$  & $77.77$ \\  
      & \cellcolor[HTML]{febafb}\lm{RoBERTa} & $17.77$  & \cellcolor[HTML]{FF8C00}$73.33$  & $75.55$ \\
       & \cellcolor[HTML]{d8d8d8}\lm{Phi4-14B} & $40.00$  & $80.00$  & \cellcolor[HTML]{FF8C00}$80.00$ \\  
       & \cellcolor[HTML]{d8d8d8}\lm{Mistral-Large} & $31.11$ & $75.55$  & $80.00$ \\  
       
       & \cellcolor[HTML]{F2CFC2}\lm{User Study}& \cellcolor[HTML]{F2CFC2}$86.67$  & \cellcolor[HTML]{F2CFC2}$73.33$  & \cellcolor[HTML]{F2CFC2}$100.00$  \\
       
       \hline

    \centering \multirow{8}{*}{\rotatebox[origin=r]{90}{\lm{Qwen2.5-14B}}} & \cellcolor[HTML]{9EC6F3}\lm{Qwen2.5-14B}     & $26.66$  & $88.88$  & $62.22$ \\
        
        & \cellcolor[HTML]{E0E6FF}\tiny{\lm{DeepSeek-R1-Distill-Qwen-14B}} & $33.33$  & \cellcolor[HTML]{C1AFEC}$93.33$  & \cellcolor[HTML]{C1AFEC}$53.33$ \\

         & \cellcolor[HTML]{FFF1D5}\lm{Qwen2.5-72B} & $37.77$  & $91.11$  & $66.66$ \\
        
       & \cellcolor[HTML]{febafb}\lm{BERT} & \cellcolor[HTML]{C1AFEC}$15.55$  & $86.66$  & $64.44$  \\  
      & \cellcolor[HTML]{febafb}\lm{RoBERTa} & $17.77$  & $86.66$  & $71.11$ \\
       & \cellcolor[HTML]{d8d8d8}\lm{Phi4-14B}& $31.11$  & $91.11$  & $60.00$  \\
       & \cellcolor[HTML]{d8d8d8}\lm{Mistral-Large}& \cellcolor[HTML]{FF8C00}$37.77$  & \cellcolor[HTML]{FF8C00}$79.13$  & \cellcolor[HTML]{FF8C00}$71.11$  \\
       
       & \cellcolor[HTML]{F2CFC2}\lm{User Study}& \cellcolor[HTML]{F2CFC2}$73.33$  & \cellcolor[HTML]{F2CFC2}$66.67$  & \cellcolor[HTML]{F2CFC2}$100.00$  \\
       
       \hline

       \centering \multirow{8}{*}{\rotatebox[origin=r]{90}{\lm{Qwen2.5-32B}}} & \cellcolor[HTML]{9EC6F3}\lm{Qwen2.5-32B}     & \cellcolor[HTML]{FF8C00}$37.77$ & $86.66$  & $57.77$  \\
        
        & \cellcolor[HTML]{E0E6FF}\tiny{\lm{DeepSeek-R1-Distill-Qwen-32B}} & $33.33$  & $88.88$  & \cellcolor[HTML]{C1AFEC}$42.22$ \\

         & \cellcolor[HTML]{FFF1D5}\lm{Qwen2.5-72B} & $35.55$  & \cellcolor[HTML]{C1AFEC}$91.11$  & $55.55$ \\
        
       & \cellcolor[HTML]{febafb}\lm{BERT} & \cellcolor[HTML]{C1AFEC}$13.33$  & $84.44$  & \cellcolor[HTML]{FF8C00}$64.44$ \\  
      
      & \cellcolor[HTML]{febafb}\lm{RoBERTa} & $20.00$ & $84.44$  & $53.33$ \\
       
       & \cellcolor[HTML]{d8d8d8}\lm{Phi4-14B} & $33.33$  & $88.88$  & $53.33$ \\ 
       
       & \cellcolor[HTML]{d8d8d8}\lm{Mistral-Large} & $31.11$  & \cellcolor[HTML]{FF8C00}$77.77$  & $51.11$ \\  
       
       & \cellcolor[HTML]{F2CFC2}\lm{User Study}& \cellcolor[HTML]{F2CFC2}$71.11$  & \cellcolor[HTML]{F2CFC2}$51.11$  & \cellcolor[HTML]{F2CFC2}$100.00$  \\
       \hline
    
       \centering \multirow{8}{*}{\rotatebox[origin=r]{90}{\lm{Llama3-70B}}} & \cellcolor[HTML]{9EC6F3}\lm{Llama3-70B} & \cellcolor[HTML]{C1AFEC}$35.55$  & \cellcolor[HTML]{C1AFEC}$91.11$  & $62.22$ \\
        
        & \cellcolor[HTML]{E0E6FF}\tiny{\lm{DeepSeek-R1-Distill-Llama-70B}} & $38.60$  & $86.66$  & \cellcolor[HTML]{C1AFEC}$46.66$ \\

         & \cellcolor[HTML]{FFF1D5}\lm{Llama3-8B} & \cellcolor[HTML]{FF8C00}$57.77$  & $86.66$  & $64.44$ \\
        
       & \cellcolor[HTML]{febafb}\lm{BERT}& $44.44$  & $86.66$  & \cellcolor[HTML]{FF8C00}$75.55$  \\
       
      & \cellcolor[HTML]{febafb}\lm{RoBERTa} & $37.77$  & $86.66$  & $60.00$ \\ 
      
      & \cellcolor[HTML]{d8d8d8}\lm{Phi4-14B} & $44.44$  & $91.11$  & $60.00$ \\  
      
       & \cellcolor[HTML]{d8d8d8}\lm{Mistral-Large} & $35.55$  & \cellcolor[HTML]{FF8C00}$75.55$  & $60.00$ \\  

        & \cellcolor[HTML]{F2CFC2}\lm{User Study}& \cellcolor[HTML]{F2CFC2}$73.33$  & \cellcolor[HTML]{F2CFC2}$62.22$  & \cellcolor[HTML]{F2CFC2}$100.00$  \\
       
        \toprule[1.5pt]     
        
        \end{tabular}
        }
        \subcaption{Counterfactual examples generated using FLARE.}
    \end{minipage}
    \caption{Label flip rate (in \%) for counterfactual examples generated by \lm{Qwen2.5-\{14B,32B\}} and \lm{Llama3-\{8B,70B\}}, evaluated across judge models exhibiting \colorbox{lightblue}{same}, \colorbox{blue}{distilled}, \colorbox{lighpurple}{same family}, and independent \colorbox{purple}{w/} and \colorbox{gray}{w/o} fine-tuning relationships, and \colorbox{lightred}{user study} on 45 selected counterfactuals (\S\ref{subsec:user_study}) each from \data{AG News}, \data{SST2} and \data{SNLI} datasets. \colorbox{deepred}{Orange-highlighted} values indicate that the judge model with the given relationship aligns \textbf{closely} with human annotators, while \colorbox{deepgreen}{purple-highlighted} values indicate the \textbf{opposite}.
    }
    \label{tab:subset_automatic_evaluation}
\end{table*}

In our user study, we randomly select 45 counterfactuals from each dataset, generated individually by five counterfactual generator models $LLM_\mathcal{G}$ (\S\ref{subsec:user_study}). To perform a sanity check and validate the representativeness of these subsets, we conduct an additional automatic evaluation on the subset of \textbf{input data} from which the 45 counterfactuals were generated. 

Table~\ref{tab:subset_automatic_evaluation} outlines the LFR performance of the generated counterfactuals across the subsets of the three dataset (\S\ref{subsec:datasets}). We observe that the entries in Table~\ref{tab:subset_automatic_evaluation} differ from those in Table~\ref{tab:automatic_evaluation}, and the optimal judge model $LLM_\mathcal{J}$ for a given counterfactual generator model $LLM_\mathcal{G}$ and dataset may vary. Nevertheless, our core finding remain consistent (\S\ref{sec:results}): judge models with an independent relationship without fine-tuning on the target dataset are the most effective at capturing actual label flips, as indicated by the alignment between automatic evaluation results and user study outcomes. In contrast, independently related judge models that have been fine-tuned on the target dataset perform the suboptimal.

\section{Counterfactual Data Augmentation}
\label{apdx:CDA}
To validate whether and to what extent counterfactual examples enhance model performance and robustness, we conduct counterfactual data augmentation (CDA) experiments using a pretrained \lm{BERT} model, $LLM_{\mathcal{C}}$, without fine-tuning on any target dataset (\S\ref{subsec:datasets}). Note that the \lm{BERT} model used for the CDA experiment as $LLM_{\mathcal{C}}$ contrasts with the \lm{BERT} model used as the judge model $LLM_\mathcal{J}$ (Table~\ref{tab:automatic_evaluation}), which is fine-tuned on the target dataset. The training set for fine-tuning the \lm{BERT} model ($LLM_{\mathcal{C}}$) consists of 500 randomly selected instances from the original training set, along with their corresponding counterfactual examples generated by the generator model ($LLM_{\mathcal{G}}$), with labels assigned by various judge models ($LLM_{\mathcal{J}}$). Our baseline is a \lm{BERT} model ($LLM_{\mathcal{B}}$), which is fine-tuned only on the same 500 randomly selected instances from the original training set. 




\subsection{Evaluation on the original test set}
\label{app:subsec:testset}
Table \ref{apx:table:augmentation} presents the accuracy of the \lm{BERT} model ($LLM_{\mathcal{C}}$) augmented with additional  counterfactual examples. Both evaluations are performed on the \textbf{original test set} of each dataset (\S\ref{subsec:datasets}). We observe that, through CDA, the performance of the \lm{BERT} model ($LLM_\mathcal{C}$) improves noticeably compared to the baseline \lm{BERT} model $LLM_\mathcal{B}$, by up to 15.13\% on average. \lm{BERT} and \lm{RoBERTa}, as $LLM_\mathcal{J}$, generally provide the most efficient labels for augmentation across the \data{AG News} and \data{SST2} datasets. This may be ascribable to the fact that the \lm{BERT} and \lm{RoBERTa} used as the judge models ($LLM_\mathcal{J}$) and the \lm{BERT} model used for CDA ($LLM_\mathcal{C}$) share the same or similar architecture, and thus, the labels provided by judge models $LLM_\mathcal{J}$ offer $LLM_\mathcal{C}$ a greater advantage compared to labels from judge models with other relationships.

Meanwhile, the performance gains observed when comparing $LLM_\mathcal{B}$ and $LLM_\mathcal{C}$, attributable to counterfactuals generated by $LLM_\mathcal{G}$, vary across tasks: \lm{Llama3-70B} generates the most effective counterfactual instances on \data{AG News} with averaged accuracy of 0.85, while \lm{Qwen2.5-7B} is the most effective on \data{SST2}, but it is the least effective on \data{SNLI}. The independent, non-fine-tuned relationship achieves the most closely aligned flip rate based on the user study. As a result, this relationship yields the best performance in counterfactual data augmentation on the \data{AG News} and \data{SST2} datasets.

\begin{table*}[ht]
    \centering
    \footnotesize
    \renewcommand*{\arraystretch}{1}
    
    \begin{minipage}{\textwidth}
    \resizebox{\columnwidth}{!}{
\begin{tabular}{ccccccccc}
\toprule[1.5pt]
 \multirow{2}{*}{ \centering  \textbf{Model}} &  \multirow{2}{*}{\centering \textbf{Judge Model}} & \multicolumn{3}{c} {\textit{\textbf{Original Test Set}}} && \multicolumn{3}{c} {\textit{\textbf{CF Set}}} \\
&& \textbf{\data{AG News}} & \textbf{\data{SST2}} & \textbf{\data{SNLI}} && \textbf{\data{AG News}} & \textbf{\data{SST2}} & \textbf{\data{SNLI}}\\
\hline
\multicolumn{2}{c}{\lm{Without CDA (Baseline)}} & 0.766 &0.779 & 0.562 && 0.307 & 0.516 & 0.289 \\
\hline



\hline

\centering \multirow{7}{*}{\rotatebox[origin=r]{90}{\lm{\tiny Llama3-8B-Instruct}}}
& \cellcolor[HTML]{9EC6F3}\lm{Meta-Llama-3-8B-Instruct} & 0.837 & 0.809 & \textbf{0.644 } && 0.267 & 0.553 & 0.311\\
& \cellcolor[HTML]{E0E6FF}\lm{DeepSeek-R1-Distill-Llama-8B} & 0.842 & 0.804 & 0.626 && \textbf{0.296} & 0.586 & 0.244\\
& \cellcolor[HTML]{FFF1D5}\lm{Meta-Llama-3-70B-Instruct} & 0.833 & 0.8337 & 0.633 && 0.278 & \textbf{0.588} & 0.276\\
& \cellcolor[HTML]{febafb}\lm{BERT} & 0.855 & 0.864 & 0.595 && 0.284 & 0.583 & 0.338\\
& \cellcolor[HTML]{febafb}\lm{RoBERTa} & \textbf{0.869} & \textbf{0.873} & 0.604 && 0.281 & 0.583 & 0.240\\
& \cellcolor[HTML]{d8d8d8}\lm{Phi4-14B} & 0.838 & 0.844 & 0.619 && 0.264 & 0.573 & \textbf{0.360}\\
& \cellcolor[HTML]{d8d8d8}\lm{Mistral-Large-Instruct} & 0.843 & 0.803 & 0.627 && 0.251 & 0.560 & 0.293\\

\hline

 \multirow{7}{*}{\rotatebox[origin=r]{90}{ \centering \lm{\tiny Qwen2.5-14B-Instruct}}} 
& \cellcolor[HTML]{9EC6F3}\lm{Qwen2.5-14B-Instruct} & 0.845 & 0.817 & 0.637 && 0.281 & 0.578 & \textbf{0.364}\\
& \cellcolor[HTML]{E0E6FF}\lm{DeepSeek-R1-Distill-Qwen-14B} & 0.837 & 0.822 & 0.621 && 0.274 & 0.546 & 0.258\\
& \cellcolor[HTML]{FFF1D5}\lm{Qwen2.5-72B-Instruct} & 0.838 & 0.811 & \textbf{0.656} && \textbf{0.291} & 0.561 & 0.280\\
& \cellcolor[HTML]{febafb}\lm{BERT} & 0.857 & \textbf{0.857} & 0.600 && 0.274 & 0.558 & 0.316\\
& \cellcolor[HTML]{febafb}\lm{RoBERTa} & \textbf{0.869} & 0.791 & 0.632 && 0.284 & 0.550 & 0.262\\
& \cellcolor[HTML]{d8d8d8}\lm{Phi4-14B} & 0.825 & 0.825 & 0.629 && 0.284 & \textbf{0.576} & 0.338\\
& \cellcolor[HTML]{d8d8d8}\lm{Mistral-Large-Instruct} & 0.803 & 0.788 & 0.642 && 0.257 & 0.542 & 0.222\\
\hline

 \multirow{7}{*}{\rotatebox[origin=r]{90}{ \tiny \centering \lm{Qwen2.5-32B-Instruct}}} 
& \cellcolor[HTML]{9EC6F3}\lm{Qwen2.5-32B-Instruct} & 0.813 & 0.836 & 0.646 && 0.277 & 0.547 & 0.347\\
& \cellcolor[HTML]{E0E6FF}\lm{DeepSeek-R1-Distill-Qwen-32B} & 0.842 & 0.836 & 0.648 && 0.286 & 0.533 & 0.351\\
& \cellcolor[HTML]{FFF1D5}\lm{Qwen2.5-72B-Instruct} & 0.774 & 0.831 & 0.646 && 0.264 & 0.532 & 0.316\\
& \cellcolor[HTML]{febafb}\lm{BERT} & 0.788 & 0.852 & 0.594 && 0.247 & 0.540 & 0.351\\
& \cellcolor[HTML]{febafb}\lm{RoBERTa} & \textbf{0.866} & \textbf{0.853} & 0.644 && 0.274 & \textbf{0.569} & 0.267\\
& \cellcolor[HTML]{d8d8d8}\lm{Phi4-14B} & 0.856 & 0.827 & \textbf{0.653} && \textbf{0.301} & 0.540 & \textbf{0.396}\\
& \cellcolor[HTML]{d8d8d8}\lm{Mistral-Large-Instruct} & 0.831 & 0.780 & 0.631&& 0.269 & 0.523 & 0.298 \\
\hline

\centering \multirow{7}{*}{\rotatebox[origin=r]{90}{\lm{\tiny Llama3-70B-Instruct}}} 

& \cellcolor[HTML]{9EC6F3}\lm{Meta-Llama-3-70B-Instruct} & 0.853 & 0.803 & 0.606 && 0.267 & 0.595 & 0.347 \\
& \cellcolor[HTML]{E0E6FF}\lm{DeepSeek-R1-Distill-Llama-70B} & 0.856 & 0.803 & 0.629 && 0.267 & 0.570 & 0.267\\
& \cellcolor[HTML]{FFF1D5}\lm{Meta-Llama-3-8B-Instruct} & 0.846 & 0.820 & 0.624 && \textbf{0.286} & 0.575 & 0.307\\
& \cellcolor[HTML]{febafb}\lm{BERT} & \textbf{0.874} & 0.820 & 0.612 && 0.254 & \textbf{0.600} & 0.324\\
& \cellcolor[HTML]{febafb}\lm{RoBERTa} & 0.853 & \textbf{0.853} & 0.629 && 0.247 & 0.595 & 0.320\\
& \cellcolor[HTML]{d8d8d8}\lm{Phi4-14B} & 0.857 & 0.812 & 0.638 && 0.267 & 0.570 & \textbf{0.369}\\
& \cellcolor[HTML]{d8d8d8}\lm{Mistral-Large-Instruct} & 0.859 & 0.791 & \textbf{0.663} && 0.235 & 0.568 & 0.289\\
\toprule[1.5pt]

\end{tabular}
}
\end{minipage}
\caption{Downstream task performance (measured in terms of accuracy) is evaluated on \textbf{\textit{test set}} and \textbf{\textit{the set of counterfactuals (out-of-distribution instances)}}  after applying counterfactual data augmentation on a \lm{BERT} model ($LLM_\mathcal{C}$). The training data consist of original examples from the target dataset, along with counterfactual examples generated by \lm{Qwen2.5-\{14B,32B\}} and \lm{Llama3-\{8B,70B\}} using \sys{FIZLE}. The counterfactual labels are provided by different judge models exhibiting various relationships: \colorbox{lightblue}{same model}, \colorbox{blue}{distilled}, \colorbox{lighpurple}{same family}, and independent models with \colorbox{purple}{fine-tuning} or \colorbox{gray}{without fine-tuning}, across the \data{AG News}, \data{SST2}, and \data{SNLI} datasets.}
\label{apx:table:augmentation}
\end{table*}

\subsection{Evaluation on the counterfactual set}
\label{apdx:subsec:cfset}
In comparison to Section~\S\ref{app:subsec:testset}, where $LLM_\mathcal{B}$ and $LLM_\mathcal{C}$ are evaluated on the test set of each dataset, we further evaluate the fine-tuned \lm{BERT} model $LLM_\mathcal{C}$ for CDA on the set of 45 selected counterfactuals, whose labels are obtained through the user study (\S\ref{subsec:user_study}). Note that the 45 selected counterfactuals and their corresponding original input texts are excluded from the training data for the model.

Table~\ref{apx:table:augmentation} illustrates the performance of the \lm{BERT} model $LLM_{\mathcal{C}}$ after fine-tuning, evaluated on the selected counterfactuals with human-annotated labels. Similar to Section~\ref{sec:results}, we count the number of instances in which $LLM_{\mathcal{J}}$ models with a specific relationship most closely or least align with human evaluation results across the three datasets. Table~\ref{apx:table:augmentation} outlines that the judge model $LLM_{\mathcal{J}}$ with an independent relationship without fine-tuning on the target dataset proves to be the most effective configuration for evaluating the validity of counterfactuals generated by $LLM_{\mathcal{G}}$, which aligns with, and further validates, our findings in Section~\ref{sec:results} (Table~\ref{tab:automatic_evaluation}). Additionally, our findings further indicate that $LLM_{\mathcal{J}}$, when configured with an independent relationship and no fine-tuning on the target dataset, generally provides LFR most closely aligned with those from human evaluation. This setup enables more effective and valid predicted labels for generated counterfactuals, which in turn contributes to better-performing and more robust models.

We calculate the Spearman correlation between the rankings of the generator and judge models in Table~\ref{tab:automatic_evaluation} and Table~\ref{apx:table:augmentation} (CF set). The results show a moderate correlation of 0.41 on the \data{AG News} dataset, indicating that the relationships might impact the performance of the CAD. Specifically, a better relationship may lead to higher accuracy on the CF test set. In contrast, a weak correlation is observed on other datasets.




\end{document}